\documentclass[runningheads]{llncs}

\PassOptionsToPackage{table}{xcolor}

\usepackage{xr-hyper}

\usepackage{eccv}

\usepackage{eccvabbrv}

\usepackage{graphicx}
\usepackage{booktabs}
\usepackage{caption}
\usepackage{multirow}
\usepackage{makecell}
\usepackage{algorithm}
\usepackage{algpseudocode}
\usepackage{pifont}
\newcommand{\cmark}{\textcolor{ForestGreen}{\ding{51}}}
\newcommand{\xmark}{\textcolor{red}{\ding{55}}}
\usepackage{color,soul} % For colored background boxes
\usepackage{wrapfig}  % in preamble
\usepackage{pgfplots}
\pgfplotsset{compat=1.18}
\usepackage{etoc}
\providecommand{\authcount}[1]{}
\etocsetlevel{title}{6}
\etocsetlevel{author}{6}

\usepackage[accsupp]{axessibility}  % Improves PDF readability for those with disabilities.
\usepackage[pagebackref,breaklinks,colorlinks,citecolor=eccvblue]{hyperref}
\usepackage[capitalize]{cleveref}
\usepackage{orcidlink}
\newsavebox{\leftblock}
\newsavebox{\rightcol}
\newsavebox{\tabblock}
\newsavebox{\figblock}

\usepackage{adjustbox}

\begin{document}

% ---------------------------------------------------------------
\title{SIGNET: Motion-Level Knowledge Transfer for Cross-Language Sign Language Translation}

% If the paper title is too long for the running head, set a short one here.
\titlerunning{SIGNET}

\author{Sobhan Asasi\orcidlink{0009-0005-6624-3341} \and
Ozge Mercanoglu Sincan\orcidlink{0000-0001-9131-0634} \and
Richard Bowden\orcidlink{0000-0003-3285-8020}}

\authorrunning{S.~Asasi et al.}

\institute{CVSSP, University of Surrey, United Kingdom\\
\ \small{\texttt{\{s.asasi, o.mercanoglusincan, r.bowden\}@surrey.ac.uk}}\\
\ \small{\href{https://cogvis-cvssp.github.io/papers/signet/}{\nolinkurl{https://cogvis-cvssp.github.io/papers/signet}}}
}

\maketitle

\begin{abstract}
    Sign language translation (SLT) remains challenging due to its high spatio-temporal complexity, long sequences, and the need to model multiple articulators without relying on gloss annotations. Existing approaches are typically tailored to individual datasets or languages and struggle to scale, while overlooking the relationships between sign languages that could inform more effective cross-lingual transfer. We present \textbf{SIGNET}, a framework that enables motion-level knowledge transfer for cross-language sign language translation.  
    Our key insight is that, although sign languages differ in grammar and lexicon, pretrained models capture motion-level visual patterns that can be reused across datasets and languages.
    \textbf{SIGNET} integrates multiple pretrained sign language backbones through an attention-based, hand-prior aggregation mechanism that guides a gated fusion network in dynamically selecting the most relevant experts.  
    Comprehensive experiments on four benchmarks (How2Sign, Phoenix14T, CSL-Daily, and MeineDGS) demonstrate state-of-the-art translation performance, and \textbf{SIGNET} also surpasses prior methods on WLASL for sign language recognition.
    \keywords{Sign Language Translation \and Cross-lingual Transfer}
\end{abstract}

\section{Introduction}
\label{sec:intro}
Sign Language Translation (SLT) aims to generate spoken or written sentences directly from sign language videos. SLT is particularly challenging because it requires the conversion of continuous video sequences with complex spatio-temporal dynamics into coherent sequences of textual tokens~\cite{nslt,sincan2025gloss}. 
This inherently multimodal problem with long temporal dependencies requires large amounts of training data and computation, limiting scalability and making it difficult for current systems to generalise beyond individual (small) datasets.
This raises a fundamental question: to what extent can motion-level knowledge learned from one sign language be reused to benefit translation in another?

To scale to larger data, recent studies have shifted towards gloss-free SLT methods~\cite{nslt,sincan2025gloss,sign2gpt,signcl,signllms}, which remove the need for intermediate gloss\footnote{Gloss is a written representation of a sign, typically denoted by one or more words from a spoken language.}  annotations, 
and directly translate from video to text. 

\begin{wrapfigure}[21]{t}{0.5\textwidth}
  \centering
   \includegraphics[width=0.5\columnwidth]{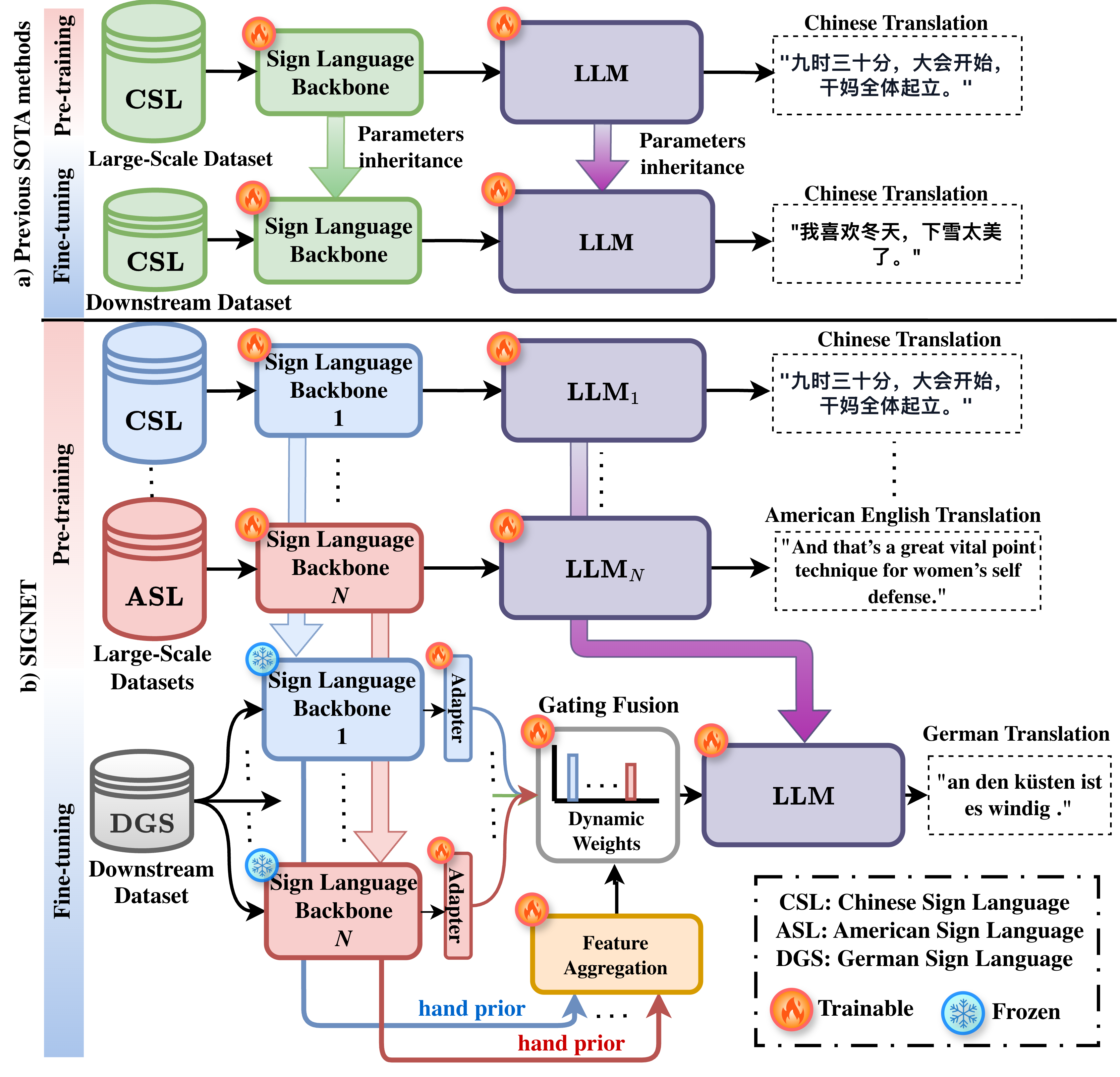}
  \caption{
Comparison between prior SLT methods and our framework. 
While existing models are language-specific and fail to generalise across other sign languages, 
\textbf{SIGNET} integrates multiple pretrained backbones via adaptive gating.
}
   \label{fig:onecol}
\end{wrapfigure}
However, most existing approaches are 
evaluated primarily on two small-scale benchmarks,
Phoenix14T~\cite{nslt} and CSL-Daily~\cite{csldaily}, which leads to overfitting and ignores the importance of  generalisation~\cite{sign2gpt,signcl,signllms,kim2025leveraging,asasibeyond}.

Most models rely on pretrained transformer- or CNN-based visual encoders that process individual frames independently, making it difficult to subsample or discard frames without losing crucial linguistic information~\cite{gfslt-vlp,guo2025bridging,sign2gpt,rust2024towards,signllms,kim2025leveraging}. 
Moreover, current methods are generally trained and fine-tuned in isolation for each dataset, discarding previously learned representations when adapting to new sign languages~\cite{gfslt-vlp,sign2gpt,signcl,signllms,guo2025bridging,kim2025leveraging,rust2024towards}. 
This fragmented training paradigm overlooks the fact that, while sign languages differ in grammar and lexicon across language families, 
they share underlying articulatory structure, and backbones pretrained on different languages can capture complementary visual cues. Leveraging such transferable representations from large-scale datasets to low-resource languages motivates our proposed method. 

As shown in Fig.~\ref{fig:onecol}, recent state-of-the-art approaches in sign language translation largely depend on extensive pretraining on language-specific datasets~\cite{rust2024towards,li2025uni,fish2025geo}. 
Although such pretraining improves performance on those languages, methods do not explore strategies to transfer the acquired knowledge to other sign languages. 
As a result, their generalisation is compromised and leads to overfitting to the domains represented in pretraining data.

To address these limitations, we propose \textbf{SIGNET} (\textbf{SI}gn Lan\textbf{G}uage \textbf{NET}\-work), a framework designed to analyse and exploit motion-level knowledge transfer across sign languages by treating pretrained sign backbones as reusable visual experts.
Our main contributions are summarised as follows:
\textbf{(\textit{i})} 
We introduce \textbf{SIGNET}, a framework that integrates independently pretrained 
sign language backbones as frozen visual experts with distinct motion-level 
expertise, combined through adaptive routing. Unlike standard Mixture-of-Experts (MoE), our 
experts encode different motion vocabularies from their pretraining languages.
\textbf{(\textit{ii})} We propose hand priors as a domain-specific routing 
signal, where hand-centric descriptors are aggregated into an input-dependent 
expert signature that guides gating, introducing articulator-level inductive 
bias absent in standard MoE routing.
\textbf{(\textit{iii})} We show that frozen cross-lingual experts can be 
effectively adapted to new sign languages through a combination of low-rank 
residual updates and contrastive alignment, requiring only 1.5M trainable 
parameters in the video encoder while matching methods that demand orders 
of magnitude more compute.
\textbf{(\textit{iv})} 
Experiments across four SLT benchmarks reveal that language-matched experts 
dominate when available, while for languages absent from pretraining the 
framework discovers complementary specialisations across experts. \textbf{SIGNET} also 
surpasses prior methods on WLASL for sign language recognition.

ASL, BSL, DGS, and CSL belong to distinct sign language lineages with limited lexical overlap and also differ in dataset domains (instructional, weather, and daily conversation). This diversity allows us to examine how motion-level priors transfer across both linguistic and domain boundaries.

\section{Related Work}
\label{sec:litrature}
\noindent\textbf{Sign Language Translation.}
SLT aims to convert continuous sign videos into spoken or written language sequences. 
Existing approaches can be broadly categorised into two paradigms: \textit{gloss-based} and \textit{gloss-free}. 
Gloss-based methods first predict an intermediate gloss sequence, representing sign-level lexical units, before translating it into text~\cite{sltunet, ts-slt,zhou2021improving,zhou2021spatial,chen2022simple,yin2021simulslt}. 
While this intermediate representation provides linguistic structure, it also introduces a dependency on expensive and labour-intensive gloss annotations, which limits data scalability and model generalisation across languages. 
To overcome these challenges, recent works have increasingly adopted \textit{gloss-free} frameworks that directly translate visual features into text, a shift largely enabled by the success of transformer architectures~\cite{gfslt-vlp,fla-llm,sign2gpt,signllms,liang2024llavasltvisuallanguagetuning}.
In this direction, several studies~\cite{gfslt-vlp,sign2gpt,fla-llm,chen2025c,asasi2025hierarchicalfeaturealignmentglossfree} employ pretext pretraining and contrastive learning to align visual and linguistic modalities. 
With the advent of large language models (LLMs), more recent approaches~\cite{sign2gpt,asasi2025hierarchicalfeaturealignmentglossfree,chen2025c, lost} leverage pretrained linguistic priors to improve translation fluency and coherence. 
Other works~\cite{guo2025bridging,kim2025leveraging,asasibeyond} incorporate LLMs or Multimodal LLMs directly into pretraining, utilising their reasoning capabilities to extract structured motion and semantic representations. 
Despite these advances, existing SLT methods remain computationally demanding and dataset-specific, with no mechanism to transfer learned representations across sign languages. 

\noindent\textbf{Efficient Sign Language Representations.} Sign language data is temporally dense, making conventional downsampling infeasible without information loss.
Although transformer- and CNN-based models~\cite{gfslt-vlp,sign2gpt,fla-llm,chen2025c,asasibeyond,kim2025leveraging,guo2025bridging} achieve strong results, their computational demands hinder large-scale or multilingual training.
Skeletal representations provide a compact alternative by efficiently encoding signer motion while maintaining linguistic structure.
Recent ST-GCN variants~\cite{stgcn,li2025uni,fish2025geo} have shown promise for sign modelling but remain limited in low-resource scenarios~\cite{ma2022learning,thatipelli2022spatio,zhang2020few}.
Our framework builds upon these efficient graph-based encoders as modular backbones and unifies them via adaptive expert fusion, achieving scalable and cross-lingual sign understanding with minimal computational overhead.

\begin{figure*}[t]
  \centering
   \includegraphics[width=\columnwidth]{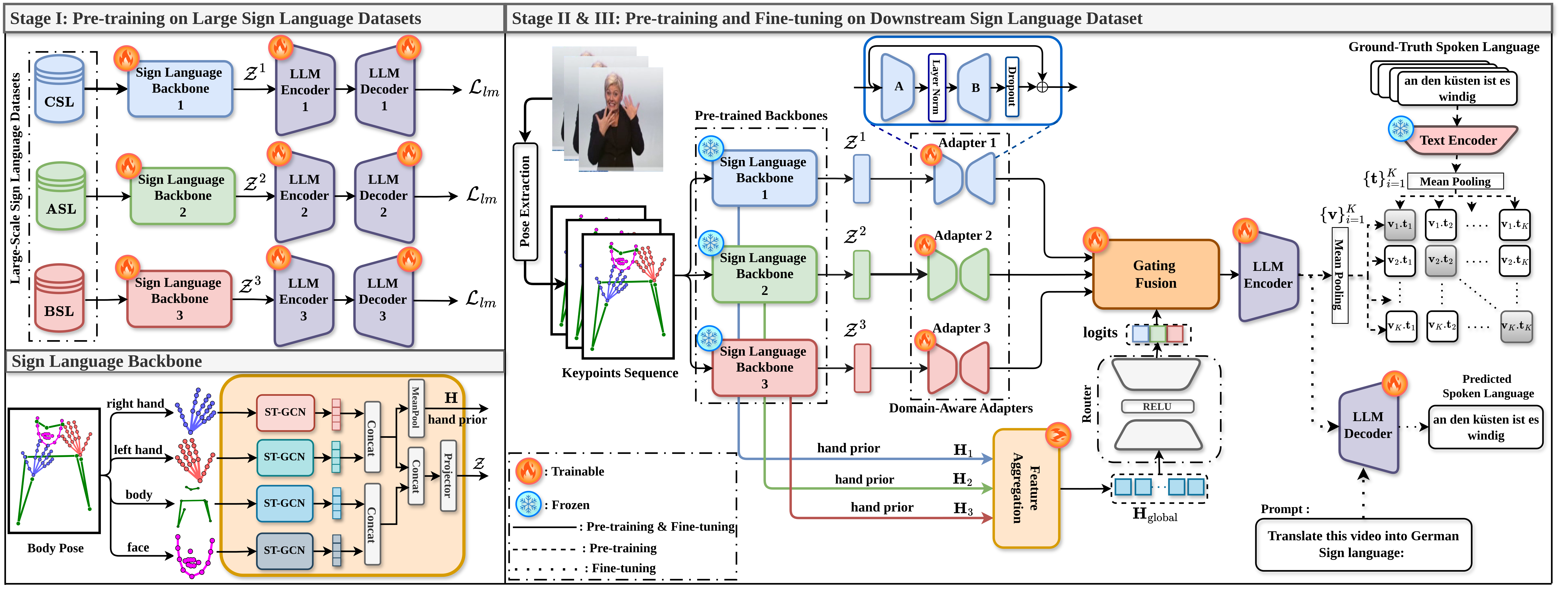}
\caption{Overview of \textbf{SIGNET}. Stage~I pretrains sign language backbones and their LLM encoder-decoders on large-scale datasets (CSL, ASL, and BSL), where each backbone (bottom left) learns part-specific motion cues via ST-GCN modules. Frozen pretrained backbones provide \emph{hand priors} to the feature aggregation module, which computes a global expert descriptor to guide the gating fusion network. Stage~II aligns visual and textual spaces via cross-modal pretraining, followed by Stage~III fine-tuning. See Fig.~\ref{fig:modules} for a detailed view of the aggregation and gating modules.}
   \label{fig:main}
\end{figure*}

\noindent\textbf{Scalability in Sign Language.}
 Acquiring large-scale aligned sign-text data remains challenging due to privacy constraints and limited accessibility. 
As a result, most prior works rely on small and domain-limited datasets such as Phoenix14T and CSL-Daily~\cite{gfslt-vlp,sign2gpt,kim2025leveraging,asasibeyond,signllms}, which often lead to overfitting. In contrast, larger and more diverse corpora such as MeineDGS~\cite{meinedgs_3} remain underexplored. 
Recent multilingual datasets~\cite{li2025uni,youtube_sl,bobsl} support scalable modelling, and pretraining-fine-tuning pipelines~\cite{li2025uni,rust2024towards} have shown improvements, yet their success depends on lexical overlap between corpora and often demands extensive resources (e.g., 64 A100 GPUs in~\cite{rust2024towards}).
Rather than assuming uniform transferability across sign languages, our framework leverages the observation that different pretrained backbones encode distinct yet complementary articulation patterns~\cite{thomas2025vallr}, which can be selectively combined to support efficient knowledge transfer across sign language domains.

\noindent\textbf{Expert Integration for Scalability.}
The mixture-of-experts (MoE) paradigm enables selective activation of relevant experts while maintaining computational efficiency~\cite{shazeer2017outrageously, fedus2022switch, du2022glam, zhou2022mixture,guo2025came,zhou2025moe}. 
Complementary developments in parameter-effici\-ent fine-tuning~\cite{houlsby2019parameter, hu2022lora, karimi2021compacter} and multimodal gating mechanisms~\cite{alayrac2022flamingo, dai2023instructblip, li2023blip} further enhance scalability.
Previous models are typically trained in isolation for each dataset and language. 
Our framework addresses this limitation by treating multiple sign language backbones, each pretrained on a distinct large-scale corpus, as visual experts.

\section{Methodology}
We first outline the training of multiple sign language backbones on large-scale datasets (Sec.~\ref{sec:preprocess}-\ref{sec:stage1}), 
then describe the modules enabling knowledge transfer to downstream datasets (Sec.~\ref{sec:adapter}-\ref{sec:gate}), 
and finally present the pretraining and fine-tuning stages for adaptation (Sec.~\ref{sec:stage2}-\ref{sec:stage3}).

\subsection{Pre-Processing}
\label{sec:preprocess}
We extract 2D skeletal keypoints from each video using RTMPose~\cite{jiang2023rtmpose} and partition them into four anatomical regions: the body, left hand, right hand, and face, denoted $\{b, lh, rh, f\}$. Each keypoint $(x, y)$ is extended with its detection confidence score to form $(x, y, c)$, where $c \in [0, 1]$. The resulting keypoint sequence for each region $r$ is denoted $\mathcal{P}_{r} \in \mathbb{R}^{T \times J_{r} \times 3}$, where $T$ is the number of frames and $J_{r}$ is the number of joints in that region.

\subsection{Sign Language Backbone}
\label{sec:backbone}
The proposed sign language backbone, denoted as $\mathcal{B}$, is designed to jointly capture motion dynamics and articulation cues across different anatomical regions of the signer (as illustrated in Fig.~\ref{fig:main}, bottom left). 
It consists of four spatio-temporal graph convolutional network (ST-GCN) modules~\cite{stgcn}, each responsible for a specific region.
%$r \in \{b, lh, rh, f\}$ corresponding to the body, left hand, right hand, and face, respectively. 
For each region, the corresponding keypoint sequence $\mathcal{P}_{r}$ is processed by its dedicated ST-GCN, $\mathcal{G}_{r}$,  within the backbone $\mathcal{B}$ to extract fine-grained spatio-temporal representations:
\begin{equation}
\mathcal{F}_{r} = \mathcal{G}_{r}(\mathcal{P}_{r}), \quad \mathcal{F}_{r} \in \mathbb{R}^{T \times C_{r}}.
\end{equation}
Formally, the overall backbone can be expressed as $\mathcal{B} = \{\mathcal{G}_{b}, \mathcal{G}_{lh}, \mathcal{G}_{rh}, \mathcal{G}_{f}\}$, where each hand ($\mathcal{G}_{rh}$,$\mathcal{G}_{lh}$) operates in parallel to encode distinct motion cues and gestures from the hands, non-manual expressions from the face ($\mathcal{G}_{f}$), and global body dynamics ($\mathcal{G}_{b}$). 
The resulting feature maps $\{\mathcal{F}_{r}\}$ are later integrated to form a comprehensive signer representation that serves as the visual foundation for translation. See Appendix for more details.

\subsection{Stage I: Pretraining on Large-Scale Datasets}
\label{sec:stage1}
As illustrated in Fig.~\ref{fig:main}, this stage establishes the foundational visual-linguistic experts used throughout the framework. 
For the $i$-th large-scale sign language dataset $\mathcal{D}^{i} = (\mathcal{P}^{i}, \mathbf{y}^{i})$,
where $\mathcal{P}^{i}$ denotes the skeletal keypoint sequences and $\mathbf{y}^{i}$ the corresponding textual translations,
a dedicated backbone $\mathcal{B}^{i}$ is jointly trained with a multilingual LLM-based encoder-decoder to align spatio-temporal visual dynamics with linguistic semantics in a shared representation space. 
Each backbone $\mathcal{B}^{i}$ comprises four region-specific ST-GCN modules 
$\{\mathcal{G}_{b}^{i}, \mathcal{G}_{lh}^{i}, \mathcal{G}_{rh}^{i}, \mathcal{G}_{f}^{i}\}$ 
that process the corresponding anatomical inputs,
$\{\mathcal{P}_{b}^{i}, \mathcal{P}_{lh}^{i}, \mathcal{P}_{rh}^{i}, \mathcal{P}_{f}^{i}\} \subset \mathcal{P}^{i}$. 
For each skeletal keypoint sequence in $\mathcal{P}^{i}$, the resulting part-level features are concatenated:
\begin{equation}
\mathcal{F}_{\operatorname{concat}}^{i} = [\mathcal{F}_{b}^{i}; \mathcal{F}_{lh}^{i}; \mathcal{F}_{rh}^{i}; \mathcal{F}_{f}^{i}] \in \mathbb{R}^{T \times C},
\end{equation}
where $C = \sum_{r} C_{r}$ is the total feature dimension across all regions. 
A linear projection maps this representation into the language model's embedding space, yielding $\mathcal{Z}^{i} \in \mathbb{R}^{T \times E}$, which is fed to the LLM encoder-decoder to predict the target sentence
$(y^{i}_1, \dots, y^{i}_{T'}) \in \mathbf{y}^{i}$ via an autoregressive decoder:
\begin{equation}
\mathcal{L}_{lm} = -\sum_{t=1}^{T'} \log P(y_t \,|\, y_{<t}, \mathcal{Z}^{i}).
\end{equation}
where $y_{t}$ represents the $t$-th token and $y_{<t}$ represents all previous tokens.

Through this end-to-end training, each backbone $\mathcal{B}^{i}$ learns to 
encode complementary spatial and temporal cues, including manual gestures 
from the hands, non-manual facial expressions, and full-body motion, into 
linguistically grounded visual representations. These pretrained backbones 
are subsequently frozen and reused as visual experts.
% , providing transferable 
% visual-linguistic priors even for low-resource sign languages.

\begin{figure*}[!t]
  \centering
   \includegraphics[width=\columnwidth]{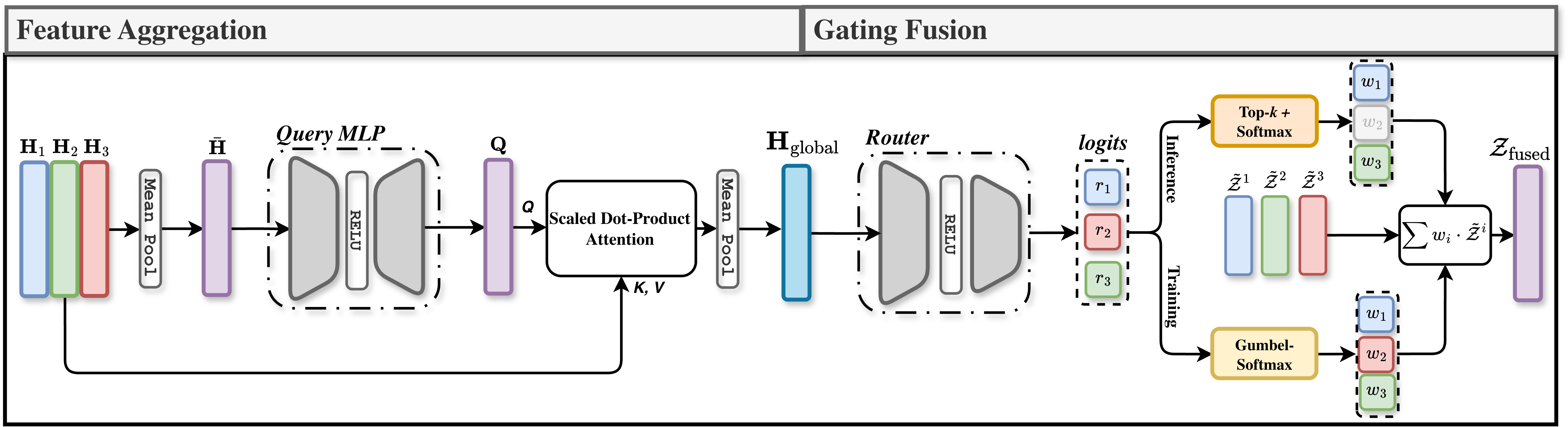}
\caption{\textbf{Feature Aggregation and Gating Fusion modules.} The feature aggregation module (left) computes a global expert descriptor $\mathbf{H}_{\text{global}}$ from hand priors via attentive pooling. The gating fusion module (right) routes $\mathbf{H}_{\text{global}}$ through a learned router to dynamically weight expert contributions, using Gumbel-Softmax during training and top-$k$ selection at inference.}
   \label{fig:modules}
\end{figure*}
\subsection{Lightweight Expert Adaptation}
\label{sec:adapter}

Given a downstream dataset
$\mathcal{D}^{*} = (\mathcal{P}^{*}, \mathbf{y}^{*})$,
where $\mathcal{P}^{*}$ denotes the skeletal input sequences
and $\mathbf{y}^{*}$ the corresponding spoken-language translations,
each input in $\mathcal{P}^{*}$ is passed through all pretrained backbones
$\{\mathcal{B}^{1}, \mathcal{B}^{2}, \dots, \mathcal{B}^{N}\}$ 
to obtain expert-level embeddings $\{\mathcal{Z}^{1}, \mathcal{Z}^{2}, \dots, \mathcal{Z}^{N}\}$ (as shown in Fig.~\ref{fig:main}) where $N$ is the number of pretrained backbones. 
While these pretrained experts provide strong visual priors, 
their projected embeddings $\mathcal{Z}^i$ may still exhibit domain mismatch 
when transferred to new datasets. 
To efficiently adapt these representations without updating the full model, 
we introduce lightweight expert adapters applied directly on top of $\mathcal{Z}^i$.
Each adapter refines the frozen expert representations through a low-rank residual transformation 
$\tilde{\mathcal{Z}}^{i} = \mathcal{Z}^{i} + (\operatorname{LN}(\mathcal{Z}^{i})\mathbf{A}_{i})\mathbf{B}_{i}$, 
where $\operatorname{LN}(\cdot)$ denotes layer normalisation, and 
$\mathbf{A}_{i} \in \mathbb{R}^{E \times R}$ and
$\mathbf{B}_{i} \in \mathbb{R}^{R \times E}$
form a learnable low-rank decomposition with rank $R \ll E$.
Following the LoRA~\cite{hu2022lora} formulation, 
the residual update $\Delta W = \mathbf{A}_{i}\mathbf{B}_{i}$ 
acts as a compact adaptation path, 
enabling the model to learn dataset-specific adjustments 
while keeping the pretrained expert parameters frozen.

\subsection{Attention-based Feature Aggregation}
\label{sec:aggregation}
To derive an input-dependent global descriptor for adaptive expert routing, 
for each pretrained backbone $\mathcal{B}^{i}$, we extract right- and left-hand features $\mathcal{F}_{rh}^{i}$ and $\mathcal{F}_{lh}^{i}$ from its corresponding ST-GCN modules, apply temporal mean pooling, and concatenate them to form the \emph{hand prior}:
\begin{equation}
\mathbf{H}_{i} = [\operatorname{Mean}_{T}(\mathcal{F}_{rh}^{i}); \operatorname{Mean}_{T}(\mathcal{F}_{lh}^{i})] \in \mathbb{R}^{B\times D}.
\end{equation}
where $B$ is the batch size and  $D = C_{rh} + C_{lh}$. 

An attention-based aggregation module (detailed in Alg.~\ref{alg:feature_aggregation}) then combines the set $[\mathbf{H}_{1}, \dots, \mathbf{H}_{N}]$ into a global descriptor $\mathbf{H}_{\text{global}}$, serving as a soft expert-selection signal that emphasises the most relevant pretrained backbones based on kinematic and semantic cues (see Fig.~\ref{fig:modules}).

\subsection{Top-\textit{k} Gating and Expert Fusion}
\label{sec:gate}
The global descriptor $\mathbf{H}_{\text{global}}$ is provided to a lightweight router network (as shown in Fig.~\ref{fig:main} and Fig.~\ref{fig:modules}), 
implemented as a two-layer MLP that outputs routing logits across the $N$ pretrained experts:
\begin{equation}
\mathbf{r} = \operatorname{MLP}(\mathbf{H}_{\text{global}}) \in \mathbb{R}^{B \times N}.
\end{equation}
During training, we employ the Gumbel-Softmax distribution~\cite{jang2016categorical,maddison2016concrete} 
to allow differentiable expert selection while maintaining stochastic exploration. 
For each expert $i$, the corresponding routing weight $w_i$ is sampled as:
\begin{equation}
\begin{aligned}
w_i &=
\frac{
\exp\!\left((r_i + g_i)/\tau_g\right)
}{
\sum_{j=1}^{N} \exp\!\left((r_j + g_j)/\tau_g\right)
}, \\
g_i &= -\log\!\left(-\log(u_i)\right), 
\quad u_i \sim \mathcal{U}(0,1).
\end{aligned}
\end{equation}
where $\tau_g$ is the temperature controlling the degree of smoothness, and
$g_i$ is the Gumbel noise drawn from a standard Gumbel distribution.
As $\tau_g \to 0$, the distribution becomes increasingly discrete,
allowing the router to approximate hard expert selection while preserving differentiability during backpropagation.
The resulting stochastic weights $\{w_i\}_{i=1}^{N}$ are used to compute a soft mixture of expert representations:
\begin{equation}
\mathcal{Z}_{\text{fused}} = \sum_{i=1}^{N} w_i \cdot \tilde{\mathcal{Z}}^{i}.
\end{equation}
At inference time, we replace the stochastic routing with a deterministic top-$k$ selection strategy.
The router selects the $k$ experts with the largest routing logits $\mathbf{r}$, 
applies a Softmax over their scores, and fuses only those $k$ experts:
\begin{equation}
\mathcal{Z}_{\operatorname{fused}} = \sum_{i \in \text{Top-}k} 
\frac{\exp(r_i)}{\sum_{j \in \text{Top-}k} \exp(r_j)} \cdot \tilde{\mathcal{Z}}^{i}.
\end{equation}
This hybrid training–inference design enables two complementary behaviours:  
(1) stochastic exploration of expert specialisation during training, 
and (2) deterministic and interpretable expert routing at test time.  

\begin{algorithm}[t]
\caption{Attention-based Feature Aggregation}
\label{alg:feature_aggregation}
\begin{algorithmic}[1]
\Require Hand priors: $\{\mathbf{H}_1, \mathbf{H}_2, \dots, \mathbf{H}_N\}$, $\mathbf{H}_i \in \mathbb{R}^{B \times D}$
\State $\mathbf{H} \gets [\mathbf{H}_1; \mathbf{H}_2; \dots; \mathbf{H}_N]$ 
\hspace{4cm}\Comment{Stack: $\mathbf{H} \in \mathbb{R}^{B \times N \times D}$}
\State $\bar{\mathbf{H}} \gets \text{MeanPool}(\mathbf{H}, \text{dim}=1)$ 
\State $\mathbf{Q} \gets \text{MLP}(\bar{\mathbf{H}})$ \hfill\Comment{Data-dependent query}
\State $\mathbf{L} \gets (\mathbf{Q}\mathbf{H}^{\top}) / \sqrt{D}$ 
\hfill\Comment{Scaled dot-product attention}
\State $\boldsymbol{\alpha} \gets \text{Softmax}(\mathbf{L}, \text{dim}=-1)$
\State $\mathbf{H}_{\text{global}} \gets \boldsymbol{\alpha} \mathbf{H}$ 
\hfill\Comment{Weighted sum over backbones}
\State $\mathbf{H}_{\text{global}} \gets \text{LayerNorm}(\mathbf{H}_{\text{global}})$
\State \Return $\mathbf{H}_{\text{global}}$
\end{algorithmic}
\end{algorithm}

\subsection{Stage II: Cross-Modal Pretraining}
\label{sec:stage2}
When adapting to a downstream sign language dataset, the training path depends on whether its spoken-language domain was present during large-scale pretraining. 
If the language has been seen, we directly reuse the corresponding pretrained LLM encoder-decoder and proceed to Stage~III. 
For languages not covered during pretraining, one of the pretrained LLM encoder-decoders in Stage I is randomly selected
and further aligned via a cross-modal contrastive pretraining stage, which bridges the gap between visual embeddings and the target linguistic space. This random choice has negligible effect on final quality (see Appendix). During this stage (as illustrated in Fig.~\ref{fig:main}, where the flow of this stage is indicated by solid and dashed lines), the sign language backbones remain frozen, 
while the adapters, routing and fusion modules, and the LLM encoder are trainable;  the LLM decoder is kept frozen to preserve its linguistic generation priors.
Let $\psi_v$ denote the LLM encoder that processes fused visual embeddings, 
$\psi_d$ the LLM decoder, and $\psi_t$ the frozen text encoder 
used for generating textual embeddings.

Given the fused visual embedding $\mathcal{Z}_{\text{fused}}$, the LLM encoder produces a global visual-language representation $\mathbf{v} = \operatorname{MeanPool}(\psi_v(\mathcal{Z}_{\text{fused}})) \in \mathbb{R}^{H}$. Similarly, for each ground-truth sentence $\mathbf{y} \in \mathbf{y}^{*}$, its textual embedding is computed using the frozen language encoder as $\mathbf{t} = \operatorname{MeanPool}(\psi_t(\mathbf{y})) \in \mathbb{R}^{H}$.
We employ a symmetric InfoNCE loss to align the visual and textual embeddings:
\begin{align}
\mathcal{L}_{\operatorname{con}} = -\frac{1}{2B} \sum_{i=1}^{B} \left[ \log \frac{\exp(\mathbf{v}_i^\top \mathbf{t}_i / \tau_c)}{\sum_{j=1}^{B} \exp(\mathbf{v}_i^\top \mathbf{t}_j / \tau_c)} + \log \frac{\exp(\mathbf{t}_i^\top \mathbf{v}_i / \tau_c)}{\sum_{j=1}^{B} \exp(\mathbf{t}_i^\top \mathbf{v}_j / \tau_c)} \right]
\end{align}
where $\tau_c$ is a learned temperature parameter controlling distribution sharpness and $B$ is the batch size.
This alignment stage grounds the visual encoder in the textual embedding space, 
enabling robust adaptation to new languages before final fine-tuning.

\subsection{Stage III: Fine-tuning on Downstream Datasets}
\label{sec:stage3}
During fine-tuning (as illustrated in Fig.~\ref{fig:main}, where the flow of this stage is indicated by solid and dotted lines), the pretrained sign language backbones remain frozen to preserve 
the robust visual-linguistic representations learned in earlier stages, 
while the adapters, attention-based fusion, gating modules, 
and the LLM encoder-decoder are updated.  
Given the skeletal keypoints extracted from each downstream video, 
the frozen backbones generate part-specific features that are aggregated 
through attention-guided fusion into the unified embedding $\mathcal{Z}_{\text{fused}}$.  
This embedding is passed to the LLM encoder, which encodes visual-semantic information,
and subsequently to the decoder $\psi_d$ to autoregressively generate the target translation.
The model is optimised using:
% a standard sequence-to-sequence objective:
\begin{equation}
\mathcal{L}_{\operatorname{fine-tune}} =
-\sum_{t=1}^{T''} \log P(y_t \,|\, y_{<t}, \mathcal{Z}_{\text{fused}}),
\end{equation}
where $\mathbf{y} = (y_1, \dots, y_{T''})\in \mathbf{y}^{*}$ denotes the target text sequence.
This stage enables efficient adaptation to new sign languages
without compromising the strong priors established during large-scale pretraining.

\section{Experiments}
\label{main:experiments}
\noindent\textbf{Large-Scale Datasets.} 
We pretrain on three large-scale, sentence-aligned video-text datasets: CSL-News~\cite{li2025uni} (1,985 hours) for Chinese Sign Language, YT-ASL~\cite{youtube_asl} (1,394 hours) for American Sign Language, and a BSL corpus constructed by combining the BSL Corpus~\cite{schembri2013building, Schembri2017British} (125 hours), the BSL subset of YT-SL25~\cite{youtube_sl} (74 hours), and BOBSL \cite{bobsl} (1,467 hours).
Each dataset trains a dedicated backbone jointly with an LLM encoder-decoder, yielding three pretrained experts ($N{=}3$) within our routing and fusion architecture.

\noindent\textbf{Downstream Datasets.} We evaluate our model on four gloss-free SLT benchmarks spanning diverse sign languages and resource levels:
How2Sign~\cite{how2sign} (79 hours, ASL) contains aligned video-text pairs of American Sign Language;
CSL-Daily~\cite{csldaily} (23 hours, CSL) features conversational Chinese Sign Language across daily topics;
Phoenix14T~\cite{nslt} (11 hours, DGS) offers German Sign Language weather broadcasts with precise alignments; and
MeineDGS~\cite{meinedgs_3} (50 hours, DGS) provides a linguistically rich corpus with high signer and vocabulary diversity.

\noindent\textbf{Metrics.} We evaluate translation performance using BLEU-1 to BLEU-4 and ROUGE-L scores, which measure n-gram precision and recall with an emphasis on longest common subsequences~\cite{bleu,rouge,brown1992class}.  Additionally, BLEURT is employed to assess semantic adequacy and fluency, capturing meaning similarity beyond lexical overlap~\cite{sellam2020bleurt}.
We report per-instance (P-I) and per-class (P-C) Top-1 accuracy for evaluating Sign Language Recognition (SLR).

\noindent\textbf{Implementation Details.} 
Our training pipeline consists of three stages: large-scale pretraining, contrastive alignment, and downstream fine-tuning. We use mT5-base as the multilingual LLM encoder-decoder across all three stages.
For Stage~I, each large-scale dataset is used to train a dedicated sign language backbone together with its LLM encoder-decoder in an end-to-end manner for 20 epochs. 
Stage~II is also trained for 20 epochs to adapt representations for unseen languages, and Stage~III runs for 40 epochs on each downstream dataset. 
% We employ the AdamW optimiser with an initial learning rate of $3\times10^{-4}$, a router-specific learning rate of $8\times10^{-4}$, and a cosine annealing scheduler across all stages. 
See Appendix for more details.

\subsection{Comparison with Prior Methods}
\label{main:results}
\begin{table*}[t]
\centering
\scriptsize
\setlength{\tabcolsep}{2.5pt}
\renewcommand{\arraystretch}{1.05}
\caption{SLT results on Phoenix14T and CSL-Daily. BLEU-1, BLEU-4, and ROUGE-L are abbreviated as B-1, B-4, and R.}
\label{tab:combined_phnx_csl}
\resizebox{\textwidth}{!}{%
\begin{tabular}{l|cc|ccc|ccc|ccc|ccc}
\toprule
\multirow{2}{*}{} 
& \multicolumn{2}{c|}{}
& \multicolumn{6}{c|}{\textbf{Phoenix14T}}
& \multicolumn{6}{c}{\textbf{CSL-Daily}} \\
\cmidrule(lr){4-9}
\cmidrule(lr){10-15}
\textbf{Method} &  \multicolumn{2}{c|}{\textbf{Modality}} 
& \multicolumn{3}{c|}{Dev}
& \multicolumn{3}{c|}{Test}
& \multicolumn{3}{c|}{Dev}
& \multicolumn{3}{c}{Test} \\
\cmidrule(lr){2-3}
\cmidrule(lr){4-6} \cmidrule(lr){7-9}
\cmidrule(lr){10-12} \cmidrule(lr){13-15}
& Pose & RGB  
& B-1\textsuperscript{$\uparrow$} & B-4\textsuperscript{$\uparrow$} & R\textsuperscript{$\uparrow$}
& B-1\textsuperscript{$\uparrow$} & B-4\textsuperscript{$\uparrow$} & R\textsuperscript{$\uparrow$}
& B-1\textsuperscript{$\uparrow$} & B-4\textsuperscript{$\uparrow$} & R\textsuperscript{$\uparrow$}
& B-1\textsuperscript{$\uparrow$} & B-4\textsuperscript{$\uparrow$} & R\textsuperscript{$\uparrow$} \\
\midrule
\rowcolor{gray!30}
\multicolumn{15}{c}{\textbf{Gloss-based}} \\
\midrule
SLTUNET~\cite{sltunet} 
&  & \ding{51}
& -- & 27.87 & 52.23
& 52.92 & 28.47 & 52.11
& -- & 23.99 & 53.58
& 54.98 & 25.01 & 54.08 \\

MMTLB~\cite{mmtlb}
&  & \ding{51}
& 53.95 & 27.61 & 53.10
& 53.97 & 28.39 & 52.65
& 53.81 & 24.42 & 53.38
& 53.31 & 23.92 & 53.25 \\

IP-SLT~\cite{ip-slt}
&  & \ding{51}
& 54.10 & 28.22 & 54.43
& 54.25 & 27.97 & 53.72
& 45.26 & 16.74 & 44.33
& 44.85 & 16.72 & 44.09 \\

TS-SLT~\cite{ts-slt}
& \ding{51} & \ding{51}
& 54.32 & 28.66 & 54.08
& 54.90 & 28.95 & 53.48
& 55.21 & 25.76 & 55.10
& 55.44 & 25.79 & 55.72 \\

\midrule
\rowcolor{gray!30}
\multicolumn{15}{c}{\textbf{Gloss-free}} \\
\midrule
GFSLT-VLP~\cite{gfslt-vlp}
&  & \ding{51}
& 44.08 & 22.12 & 43.72
& 43.71 & 21.44 & 42.49
& 39.20 & 11.07 & 36.70
& 39.37 & 11.00 & 36.44 \\

FLa-LLM~\cite{fla-llm}
&  & \ding{51}
& -- & -- & --
& 46.29 & 23.09 & 45.27
& -- & -- & --
& 37.13 & 14.20 & 37.25 \\

Sign2GPT~\cite{sign2gpt}
&  & \ding{51}
& -- & -- & --
& 49.54 & 22.52 & 48.90
& -- & -- & --
& 41.75 & 15.40 & 42.36 \\

SignLLM~\cite{signllms}
&  & \ding{51}
& 46.88 & 25.25 & 47.23
& 45.21 & 23.40 & 44.49
& 42.45 & 12.23 & 39.18
& 39.55 & 15.75 & 39.91 \\

MMSLT~\cite{kim2025leveraging}
&  & \ding{51}
& 48.73 & 25.47 & 48.58
& 48.92 & 25.73 & 47.97
& 50.05 & 20.51 & 48.53
& 49.87 & 21.11 & 48.92 \\

BeyondGloss~\cite{asasibeyond}
&  & \ding{51} 
& -- & -- & --
& 52.38 & 25.49 & 52.89
& -- & -- & --
& 53.12 & 21.53 & 53.46 \\

PGG-SLT~\cite{guo2025bridging}
&  & \ding{51}
& 53.84 & 27.53 & 52.85
& 54.02 & 27.32 & 52.56
& -- & -- & --
& -- & -- & -- \\

Uni-Sign~\cite{li2025uni}
& \ding{51} & \ding{51}
& -- & -- & --
& -- & -- & --
& 55.30 & 26.25 & 56.03
& 55.08 & 26.36 & 56.51 \\

Geo-Sign~\cite{fish2025geo}
& \ding{51} & 
& -- & -- & --
& -- & -- & --
& 55.57 & 27.05 & 57.27
& 55.89 & 27.42 & 57.95 \\

\midrule
\rowcolor[HTML]{e3f2fd}
\textbf{SIGNET}
& \ding{51} & 
& \textbf{54.15} & \textbf{27.78} & \textbf{53.01}
& \textbf{54.10} & \textbf{27.82} & \textbf{53.05}
& \textbf{56.90} & \textbf{28.27} & \textbf{58.48}
& \textbf{56.77} & \textbf{28.51} & \textbf{58.66} \\
\bottomrule
\end{tabular}}
\end{table*}
\textbf{Results on Phoenix14T.} As seen in Tab.~\ref{tab:combined_phnx_csl}, \textbf{SIGNET} achieves state-of-the-art results among gloss-free methods using only skeletal keypoints, without requiring RGB frames or gloss annotations. Compared to  PGG-SLT~\cite{guo2025bridging}, which leverages an LLM to predict and order glosses through prompt-based reasoning, \textbf{SIGNET} achieves a +0.50 BLEU-4 gain with a substantially lighter input representation, highlighting the effectiveness of multi-expert fusion for cross-lingual transfer.

\noindent\textbf{Results on CSL-Daily.} Based on Tab.~\ref{tab:combined_phnx_csl}, on this dataset, our method achieves clear gains over prior approaches, 
outperforming RGB-based models by +6.9 BLEU-4 and showing consistent improvements
over pose-based methods, with a +1.09 BLEU-4 increase compared to Geo-Sign~\cite{fish2025geo}. 
While Geo-Sign employs computationally intensive geometric operations that introduce higher latency during training~\cite{fish2025geo}, 
our framework attains superior efficiency through a light\-weight adaptation mechanism that scales across multiple pretrained backbones. 

\noindent\textbf{Results on How2Sign.} 
Tab.~\ref{tab:how2sign} highlights that our model achieves state-of-the-art performance on How2Sign, 
outperforming the pose-based Uni-Sign~\cite{li2025uni} and Geo-Sign~\cite{fish2025geo}, 
and comparable results to the RGB-based SSVP-SLT-LSP~\cite{rust2024towards}. 
However, this method requires an extremely demanding training setup, 
utilising 64 A100 GPUs for two weeks and leveraging both YT-ASL and How2Sign during pretraining as well as during fine-tuning, making it difficult to extend to additional datasets~\cite{rust2024towards}. 
In contrast, our approach achieves these results through an efficient and scalable design that only requires fine-tuning on How2Sign.

\noindent\textbf{Results on MeineDGS.} 
Tab.~\ref{tab:meindgs} shows that \textbf{SIGNET} generalises to MeineDGS, a dataset whose language (DGS) was not seen during pretraining, showcasing the transferability of motion-level priors to new sign languages. Unlike Sincan et al.~\cite{sincan2025spotter+}, which partially leverage ground-truth glosses and GPT, our method operates in a fully gloss-free manner while achieving better performance.

\begin{table*}[!t]
\centering

% ---------- LEFT SIDE (How2Sign) — measured into a box ----------
\sbox{\leftblock}{%
\begin{minipage}[t]{0.52\textwidth}
\centering
\scriptsize
\setlength{\tabcolsep}{3pt}
\renewcommand{\arraystretch}{1}
\caption{SLT results on How2Sign. $\dagger$ denotes pretraining on YT-ASL and How2Sign. BLEU-1, BLEU-4, ROUGE-L and BLEURT are abbreviated as B-1, B-4, R and B-RT.}
\label{tab:how2sign}
\scalebox{0.82}{%
\begin{tabular}{l|cc|cccc}
\toprule
\multirow{2}{*}{\textbf{Method}} & \multicolumn{2}{c|}{\textbf{Modality}} & \multicolumn{4}{c}{\textbf{Test}} \\
\cmidrule(lr){2-3} \cmidrule(lr){4-7}
 & Pose & RGB & B-1\textsuperscript{$\uparrow$} & B-4\textsuperscript{$\uparrow$} & R\textsuperscript{$\uparrow$} & B-RT\textsuperscript{$\uparrow$} \\
\midrule
\rowcolor{gray!30}
\multicolumn{7}{c}{\textbf{Gloss-free}} \\
\midrule
\color{gray}
SSVP-SLT-LSP$\dagger$~\cite{rust2024towards} &  & \color{gray}\ding{51} & \color{gray}43.2 & \color{gray}15.5 & \color{gray}38.4 & \color{gray}49.6 \\
\midrule
YouTube-ASL~\cite{youtube_asl} & \ding{51} &  & 37.8 & 12.4 & \textbf{--} & 46.6 \\
SLT-IV~\cite{slt-iv} &  & \ding{51} & 34.0 & 8.0 & \textbf{--} & \textbf{--} \\
C$^2$RL~\cite{chen2025c} &  & \ding{51} & 29.1 & 9.4 & 27.0 & \textbf{--} \\
FLa-LLM~\cite{fla-llm} &  & \ding{51} & 29.8 & 9.7 & 27.8 & \textbf{--} \\
PGG-SLT~\cite{guo2025bridging} &  & \ding{51} & 40.8 & 13.7 &  32.9 & \textbf{--} \\
Jang et al.~\cite{lost} &  & \ding{51} & \textbf{--} & 12.7 & 32.5 & 45.3 \\
SignMusketeers~\cite{gueuwou2025signmusketeers} &  & \ding{51} & 41.5 & 14.3 & \textbf{--} & \textbf{--} \\
SSVP-SLT~\cite{rust2024towards} &  & \ding{51} & 41.9 & 14.7 & 37.8 & 49.3 \\
Uni-Sign~\cite{li2025uni} & \ding{51} & \ding{51} & 40.2 & 14.9 & 36.0 & 49.4 \\
Geo-Sign~\cite{fish2025geo} & \ding{51} &  & 40.8 & 15.1 & 35.4 & \textbf{--} \\
\midrule
\rowcolor[HTML]{e3f2fd}
\textbf{SIGNET} & \ding{51} &  & \textbf{41.1} & \textbf{15.4} & \textbf{36.4} & \textbf{49.5} \\
\bottomrule
\end{tabular}}
\end{minipage}%
}

% ---------- TYPESET both columns at equal height ----------
\usebox{\leftblock}%
\hfill
\begin{minipage}[t][\dimexpr\ht\leftblock+\dp\leftblock\relax][t]{0.45\textwidth}

  % ---- MeineDGS (top right) ----
  \centering
  \footnotesize
  \setlength{\tabcolsep}{4pt}
  \renewcommand{\arraystretch}{1.05}
  \caption{SLT results on MeineDGS.}
  \label{tab:meindgs}
  \resizebox{\linewidth}{!}{%
  \begin{tabular}{l|cc|cccc}
  \toprule
  \multirow{2}{*}{\textbf{Method}} & \multicolumn{2}{c|}{\textbf{Modality}}  & \multicolumn{4}{c}{\textbf{Test}} \\ \cmidrule(lr){2-3} \cmidrule(lr){4-7}
   & Pose & RGB & B-1\textsuperscript{$\uparrow$} & B-4\textsuperscript{$\uparrow$} & R\textsuperscript{$\uparrow$} & B-RT\textsuperscript{$\uparrow$} \\
  \midrule
  \rowcolor{gray!30}
  \multicolumn{7}{c}{\textbf{Gloss-free}} \\
  \midrule
  Spotter+GPT~\cite{sincan2025spotter+} &  & \ding{51} & 14.82 & 0.64 & -- & 21.62 \\
  Spotter+Transformer~\cite{sincan2025spotter+} &  & \ding{51} & 19.50 & 1.08 & -- & 19.01 \\
  Sub-GT+GPT~\cite{sincan2025spotter+} &  & \ding{51} & 16.65 & 1.55 & -- & 29.72 \\
  \midrule
  \rowcolor[HTML]{e3f2fd}
  \textbf{SIGNET} & \ding{51} &  & \textbf{24.20} & \textbf{2.75} & \textbf{13.6} & \textbf{33.58} \\
  \bottomrule
  \end{tabular}}

  \vfill   % <-- collects all the slack here, pushing WLASL to the bottom

  % ---- WLASL (bottom right) ----
  \centering
  \footnotesize
  \setlength{\tabcolsep}{4pt}
  \renewcommand{\arraystretch}{1.1}
  \caption{SLR results on WLASL. P-I and P-C denote per-instance and per-class Top-1 accuracy, respectively.}
  \label{tab:wlasl}
  \resizebox{\linewidth}{!}{%
  \begin{tabular}{l|cc|cc|cc}
  \hline
  \multirow{2}{*}{\textbf{Method}} & \multicolumn{2}{c|}{\textbf{Modality}} & \multicolumn{2}{c|}{\textbf{WLASL100}} & \multicolumn{2}{c}{\textbf{WLASL2000}} \\ \cmidrule(lr){2-3} \cmidrule(lr){4-5} \cmidrule(lr){6-7}
   & Pose & RGB & P-I\textsuperscript{$\uparrow$} & P-C\textsuperscript{$\uparrow$} & P-I\textsuperscript{$\uparrow$} & P-C\textsuperscript{$\uparrow$} \\
  \hline
  NLA-SLR~\cite{NLA-SLR} & \ding{51} & \ding{51} & 91.47 & 92.17 & 61.05 & 58.05 \\
  SignRep~\cite{wong2025signrep} &  & \ding{51} & \textbf{--} & \textbf{--} & 61.05 & 58.89 \\
  Uni-Sign~\cite{li2025uni} & \ding{51} & \ding{51} & 92.25 & 92.67 & 63.52 & 61.32 \\
  Geo-Sign~\cite{fish2025geo} & \ding{51} &  & \textbf{--} & \textbf{--} & 63.64 & 61.89 \\
  \midrule
  \rowcolor[HTML]{e3f2fd}
  \textbf{SIGNET} & \ding{51} &  & \textbf{93.26} & \textbf{93.47} & \textbf{64.87} & \textbf{62.07} \\
  \hline
  \end{tabular}}

\end{minipage}
\end{table*}

\noindent\textbf{Results on WLASL.}
We evaluate \textbf{SIGNET} for isolated sign recognition on WLASL100 and WLASL2000~\cite{wlasl} to assess the transferability of learned visual-linguistic features. Without model modifications, we skip Stage II and replace target sentences with individual glosses in Stage III training. As shown in Tab.~\ref{tab:wlasl}, \textbf{SIGNET} outperforms existing pose-based and RGB-based methods, suggesting that representations learned by \textbf{SIGNET} effectively capture transferable sign semantics beyond translation tasks. 

\noindent\textbf{Computational Efficiency.}
Tab.~\ref{tab:efficiency} compares parameter counts and forward latency across methods. Despite using multiple pretrained backbones, \textbf{SIGNET} requires only 1.5M trainable parameters in the video encoder, as backbone features can be pre-computed and cached. Stage~I pretraining is a one-time cost (576 GPU-hours, 4$\times$RTX 3090); downstream adaptation requires only ${\sim}$24 GPU-hours on a single RTX 3090. In comparison, SSVP-SLT-LSP demands 21,504 A100-hours for a single language, yet \textbf{SIGNET} matches its How2Sign performance (15.4 vs. 15.5 B-4) using 64$\times$ fewer GPUs without requiring How2Sign during pretraining. \textbf{SIGNET} achieves a per-batch training latency of 320\,ms, compared to Uni-Sign (416\,ms) and Geo-Sign (2550\,ms), both of which train and evaluate on a single language and do not address cross-lingual transfer.

\noindent\textbf{Qualitative Results.} See Appendix for qualitative examples.\\

\begin{table}[t]
\centering
\scriptsize
\setlength{\tabcolsep}{1.5pt}
\renewcommand{\arraystretch}{1.07}
% Measure the table into a box so the caption can be set to the same width.
\sbox0{%
\begin{tabular}{l|cc|cc|cc|c}
\toprule
\multirow{2}{*}{\textbf{Method}} & \multicolumn{2}{c|}{\textbf{Video Encoder}} & \multicolumn{2}{c|}{\textbf{Language Model}} & \multicolumn{3}{c}{\textbf{Full Model}} \\
\cline{2-8}
  & \#Total & \#Train  & \#Total & \#Train & \#Total & \#Train & Fwd Latency \\
\midrule
\rowcolor{gray!30}
\multicolumn{8}{c}{\textbf{RGB-based Methods}} \\ \midrule
Sign2GPT~\cite{sign2gpt}  & 34.9M & 12.8M  & 1736.7M & 3.8M & 1771.6M & 16.7M & -- \\
C$^2$RL~\cite{chen2025c} & 11.7M & 11.7M  & 680M & 680M & 691.7M & 691.7M & -- \\
\midrule
\rowcolor{gray!30}
\multicolumn{8}{c}{\textbf{Skeleton-based Methods}} \\ \midrule
Uni-Sign~\cite{li2025uni}  & 9.7M & 9.7M & 582.4M & 582.4M & 592.1M & 592.1M & 416 ms \\
Geo-Sign~\cite{fish2025geo} & 6.7M & 6.7M  & 582.4M & 582.4M & 589.1M & 589.1M &  2550 ms \\
\midrule
\rowcolor[HTML]{e3f2fd}
\textbf{SIGNET}  & 17.4M$^{\dagger}$ & \textbf{1.5M} & 582.4M & 582.4M & 599.8M & 583.9M & \textbf{320 ms} \\
\bottomrule
\end{tabular}%
}%
\begin{minipage}{\wd0}
\centering
\caption{Computational efficiency comparison. Parameters in millions (M); forward latency in milliseconds (ms) per training batch (batch size 8). $^{\dagger}$Backbone features can be pre-computed and cached after Stage~I.
The reported latency is \emph{non-cached} (all three backbones run per batch).
}
\label{tab:efficiency}
\usebox0
\end{minipage}
\end{table}

\subsection{Ablation Study}
\label{main:ablation}
\textbf{Cross-lingual Transfer Benefits.}
Tab.~\ref{tab:frozen_backbone} compares performance when fine-tuning with trainable versus frozen backbones pretrained on large-scale datasets.
For this evaluation, we pretrain a sign language backbone and an LLM on a large-scale dataset (similar to Sec.~\ref{sec:stage1}) and without adding any extra modules, we fine-tune it directly on the downstream datasets after loading the pretrained weights.
This table confirms the assumptions that our paper is based upon and that adapting frozen backbones yield substantially higher performance across cross-lingual settings,
while trainable backbones suffer from overfitting and catastrophic forgetting.
Notably, when the pretraining and downstream datasets share the same language, both settings perform comparably.
However, in cross-lingual transfer (e.g., CSL-News~$\rightarrow$~Phoenix14T or YT-ASL~$\rightarrow$~CSL-Daily), the frozen backbone consistently achieves stronger results, with some scores comparable to prior methods.
These findings suggest that pretrained backbones retain transferable articulatory representations across sign languages, supporting our design of keeping them frozen and adapting through lightweight modules.

\noindent\textbf{Feature Aggregation Input.} In Tab.~\ref{ab:tab_2}, we observe that using only the hand features as input priors to the feature aggregation module yields the best result across all datasets, while incorporating body and facial priors reduces translation quality. We attribute this to the attention-based aggregation being optimised for compact, semantically dense inputs; the higher-dimensional body and face features likely introduce noise into the routing signal. This is consistent with the role of hands as the primary carriers of lexical and semantic content in sign languages, making them the most effective cue to distinguish expert contributions.

\noindent\textbf{Components Contributions.} Tab.~\ref{tab:contribution} presents the effect of individual components and backbone configurations 
on translation performance.
Specifically, removing Gating replaces learned routing with uniform averaging, removing Adapter discards the low-rank residual updates, and removing Contrastive skips Stage~II alignment.
We systematically evaluate the contribution of the \textit{Adapter}, \textit{Gating}, and \textit{Contrastive} modules 
under different numbers of pretrained backbones ($N \in \{1,2,3\}$) and routing top-$k$ selections.  
On Phoenix14T, the best performance is achieved when using
three backbones with $k=2$ and all components enabled, demonstrating that 
multi-expert fusion and contrastive loss jointly enhance cross-modal correspondence.  
Interestingly, for CSL-Daily, removing the contrastive loss slightly improves results, 
suggesting that excessive cross-lingual regularisation may interfere with datasets 
that already share language overlap with pretraining corpora.  
Overall, the results highlight the complementary role of each component, 
\textit{Adapters} facilitate efficient transfer, 
\textit{Gating} improves dynamic expert selection, 
and \textit{Contrastive alignment} strengthens visual-linguistic grounding when domain or language gaps are present. 
With three pretrained backbones, several combinations are possible when selecting one or two experts. The reported results correspond to the best configuration. 

\begin{table*}[t]
\centering

% ---------- RIGHT SIDE (reference) measured into a box ----------
% Pretend the two left tables already passed, so the caption inside
% the box bakes the correct (third) number; then fully restore the counter.
\addtocounter{table}{2}%
\sbox{\rightcol}{%
\begin{minipage}[t]{0.52\textwidth}
\centering
\footnotesize
\setlength{\tabcolsep}{3pt}
\renewcommand{\arraystretch}{0.9}

\captionsetup{type=table,width=\linewidth,singlelinecheck=false}
\captionof{table}{Ablation on top-$k$ gating, components, and backbone count.
\textbf{(i)} No Gating replaces learned routing weights with uniform averaging over experts; \textbf{(ii)} No Adapter removes the low-rank residual updates while keeping frozen expert features; \textbf{(iii)} No Contrastive skips Stage II alignment and proceeds directly to downstream fine-tuning.}
\label{tab:contribution}

\resizebox{\linewidth}{!}{%
\begin{tabular}{r|ccc|cc|cc}
\toprule
\multirow{2}{*}{$k$} & \multicolumn{3}{c|}{\textbf{Components}} & \multicolumn{2}{c|}{\textbf{Phoenix14T}} & \multicolumn{2}{c}{\textbf{CSL-Daily}} \\ \cmidrule(lr){2-4} \cmidrule(lr){5-6} \cmidrule{7-8}
 & Adapter & Gating & Contrastive & B-4\textsuperscript{$\uparrow$} & R\textsuperscript{$\uparrow$} & B-4\textsuperscript{$\uparrow$} & R\textsuperscript{$\uparrow$} \\
\midrule
\rowcolor{gray!30}
\multicolumn{8}{c}{\textbf{1 Backbone}} \\
\midrule
\textbf{--} & \xmark & \xmark & \xmark & 24.05 & 47.52 & 25.56 & 53.82 \\
\textbf{--} & \cmark & \xmark & \xmark & 23.85 & 46.60 & 26.23 & 55.45 \\
\textbf{--} & \xmark & \xmark & \cmark & 24.65 & 46.62 & 26.20 & 54.44 \\
\textbf{--} & \cmark & \xmark & \cmark & 25.12 & 47.78 & 25.58 & 53.95 \\
\midrule
\rowcolor{gray!30}
\multicolumn{8}{c}{\textbf{2 Backbones}} \\
\midrule
1 & \xmark & \cmark & \xmark & 25.36 & 47.88 & 24.45 & 53.92 \\
1 & \cmark & \cmark & \xmark & 26.35 & 49.48 & 27.30 & 56.12 \\
1 & \cmark & \cmark & \cmark & 27.15 & 51.08 & 26.65 & 55.12 \\
2 & \cmark & \cmark & \cmark & 27.10 & 51.04 & 26.10 & 54.23 \\
\midrule
\rowcolor{gray!30}
\multicolumn{8}{c}{\textbf{3 Backbones}} \\
\midrule
1 & \cmark & \cmark & \cmark & 27.20 & 50.98 & 27.25 & 55.62 \\
\rowcolor[HTML]{e3f2fd}
\textbf{2} & \cmark & \cmark & \cmark & \textbf{27.78} & \textbf{53.01} & 26.50 & 55.84 \\
\rowcolor[HTML]{e3f2fd}
\textbf{2} & \cmark & \cmark & \xmark & 27.63 & 52.87 & \textbf{28.27} & \textbf{58.48} \\
3 & \cmark & \cmark & \cmark & 26.30 & 49.54 & 26.90 & 55.52 \\
\bottomrule
\end{tabular}}
\end{minipage}%
}%
\addtocounter{table}{-3}% undo the +2 and the +1 the caption added; counter back to start

% ---------- TYPESET both columns at equal height ----------
% LEFT SIDE (stacked), stretched to match the right column.
\begin{minipage}[t][\dimexpr\ht\rightcol+\dp\rightcol\relax][t]{0.45\textwidth}

  % ---- Left Top (Frozen Backbone) ----
  \centering
  \footnotesize
  \setlength{\tabcolsep}{5pt}
  \renewcommand{\arraystretch}{1}

  \captionsetup{type=table,width=\linewidth,singlelinecheck=false}
  \captionof{table}{Adaptation impact on addressing overfitting. \textcolor{ForestGreen}{$\uparrow$}/\textcolor{red}{$\downarrow$}: B-4 change from freezing.}
  \label{tab:frozen_backbone}

  \resizebox{\linewidth}{!}{%
  \begin{tabular}{c|c|c|c}
  \toprule
  \multirow{2}{*}{\makecell{\textbf{Pretraining}\\ \textbf{Dataset}}} & \textbf{Phoenix14T} & \textbf{CSL-Daily} & \textbf{How2Sign} \\ \cmidrule(lr){2-4}
   & B-4\textsuperscript{$\uparrow$} & B-4\textsuperscript{$\uparrow$} & B-4\textsuperscript{$\uparrow$} \\
  \midrule
  \rowcolor{gray!30}
  \multicolumn{4}{c}{\textbf{Trainable Backbone}} \\
  \midrule
  CSL-News & 11.47 & \textbf{25.56} & 6.8 \\
  YT-ASL & 12.04 & 13.8 & \textbf{14.4} \\
  \midrule
  \rowcolor{gray!30}
  \multicolumn{4}{c}{\textbf{Frozen Backbone}} \\
  \midrule
  \rowcolor[HTML]{e3f2fd}
  CSL-News & \textbf{23.30}\,{\scriptsize\textcolor{ForestGreen}{$\uparrow$\textbf{11.8}}} & 24.65\,{\scriptsize\textcolor{red}{$\downarrow$\textbf{0.9}}} & \textbf{10.3}\,{\scriptsize\textcolor{ForestGreen}{$\uparrow$\textbf{3.5}}} \\
  \rowcolor[HTML]{e3f2fd}
  YT-ASL & \textbf{24.05}\,{\scriptsize\textcolor{ForestGreen}{$\uparrow$\textbf{12.0}}} & \textbf{20.7}\,{\scriptsize\textcolor{ForestGreen}{$\uparrow$\textbf{6.9}}} & 13.1\,{\scriptsize\textcolor{red}{$\downarrow$\textbf{1.3}}} \\
  \bottomrule
  \end{tabular}}

  \vfill   % <-- collects the slack here, pushing the lower table to the bottom

  % ---- Left Bottom (Input Configurations) ----
  \centering
  \footnotesize
  \setlength{\tabcolsep}{5pt}
  \renewcommand{\arraystretch}{0.8}

  \captionsetup{type=table,width=\linewidth,singlelinecheck=false}
  \captionof{table}{Impact of different input configurations for feature aggregation on performance.}
  \label{ab:tab_2}

  \resizebox{\linewidth}{!}{%
  \begin{tabular}{ccc|ccc}
  \toprule
  \multicolumn{3}{c|}{\textbf{Phoenix14T}} & \multicolumn{3}{c}{\textbf{CSL-Daily}} \\ \cmidrule(lr){1-3} \cmidrule(lr){4-6}
  B-1\textsuperscript{$\uparrow$} & B-4\textsuperscript{$\uparrow$} & R\textsuperscript{$\uparrow$} & B-1\textsuperscript{$\uparrow$} & B-4\textsuperscript{$\uparrow$} & R\textsuperscript{$\uparrow$} \\
  \midrule
  \rowcolor{gray!30}
  \multicolumn{6}{c}{\textbf{Hands}} \\
  \midrule
  \rowcolor[HTML]{e3f2fd}
  \textbf{54.15} & \textbf{27.78} & \textbf{53.01} & \textbf{56.90} & \textbf{28.27} & \textbf{58.48} \\
  \midrule
  \rowcolor{gray!30}
  \multicolumn{6}{c}{\textbf{Hands + Body}} \\
  \midrule
  53.83 & 26.23 & 50.81 & 54.67 & 26.78 & 56.45 \\
  \midrule
  \rowcolor{gray!30}
  \multicolumn{6}{c}{\textbf{Hands + Body + Face}} \\
  \midrule
  51.11 & 25.12 & 48.08 & 52.56 & 24.94 & 53.64 \\
  \bottomrule
  \end{tabular}}

\end{minipage}%
\hfill
\usebox{\rightcol}
\addtocounter{table}{1}% counter ended at the left tables (7); advance to 8 (contribution) so later tables continue from 9
\end{table*}

\noindent\textbf{Effect of $k$ on SLT.}
According to Tab.~\ref{tab:contribution}, performance does not increase monotonically with $k$. The best results arise with three backbones and $k{=}2$. Using a single expert ($k{=}1$) limits cues, while using all experts ($k{=}3$) introduces redundancy and over-smoothing. $k{=}2$ strikes a sparsity--diversity trade-off, combining two complementary experts while keeping the mixture selective. Empirically, this setting yields higher gate confidence and lower pairwise expert redundancy.

\begin{figure}[t]
  % ---------- LEFT: table, caption on top ----------
  \begin{minipage}[t]{0.56\textwidth}
    \null
    \centering
    \footnotesize
    \setlength{\tabcolsep}{2pt}
    \renewcommand{\arraystretch}{1.1}

    \captionsetup{type=table,width=\linewidth,singlelinecheck=false}
    \captionof{table}{Leave-one-expert-out analysis ($k{=}2$). $\Delta$ denotes the BLEU-4 drop when each pretrained backbone is removed from the full three-expert model; the largest drop per dataset is \textbf{bolded}. Per dataset, \colorbox{ForestGreen!45}{green}/\colorbox{red!45}{red} shade higher BLEU-4\,/\,larger drop.}
    \label{tab:leave-one-out}

    \resizebox{\linewidth}{!}{%
    \begin{tabular}{l|cc|cc|cc|cc}
% \toprule
 & \multicolumn{2}{c|}{\textbf{Phoenix14T}} & \multicolumn{2}{c|}{\textbf{CSL-Daily}} & \multicolumn{2}{c|}{\textbf{How2Sign}} & \multicolumn{2}{c}{\textbf{MeineDGS}} \\
 \cmidrule(lr){2-9}
 & B-4$\uparrow$ & $\Delta$ & B-4$\uparrow$ & $\Delta$ & B-4$\uparrow$ & $\Delta$ & B-4$\uparrow$ & $\Delta$ \\
\midrule
\rowcolor[HTML]{e3f2fd}
\textbf{SIGNET} & \textbf{27.78} & -- & \textbf{28.27} & -- & \textbf{15.4} & -- & \textbf{2.75} & -- \\
\midrule
w/o \textbf{ASL expert} & \cellcolor{ForestGreen!10}23.04 & \cellcolor{red!60}\textbf{4.74} & \cellcolor{ForestGreen!40}26.79 & \cellcolor{red!20}1.48 & \cellcolor{ForestGreen!10}9.8 & \cellcolor{red!60}\textbf{5.6} & \cellcolor{ForestGreen!10}1.05 & \cellcolor{red!60}\textbf{1.7} \\
w/o \textbf{BSL expert} & \cellcolor{ForestGreen!32}24.02 & \cellcolor{red!30}3.76 & \cellcolor{ForestGreen!48}27.84 & \cellcolor{red!10}0.43 & \cellcolor{ForestGreen!48}14.9 & \cellcolor{red!10}0.5 & \cellcolor{ForestGreen!32}1.65 & \cellcolor{red!30}1.1\\
w/o \textbf{CSL expert} & \cellcolor{ForestGreen!48}24.68 & \cellcolor{red!10}3.10 & \cellcolor{ForestGreen!10}21.71 & \cellcolor{red!60}\textbf{6.56} & \cellcolor{ForestGreen!40}14.1 & \cellcolor{red!20}1.3 & \cellcolor{ForestGreen!48}2.05 & \cellcolor{red!10}0.7 \\
\bottomrule
\end{tabular}%
    }
  \end{minipage}%
  \hfill
  % ---------- RIGHT: figure, caption below ----------
  \begin{minipage}[t]{0.4\textwidth}
    \null
    \centering
    \includegraphics[width=\linewidth]{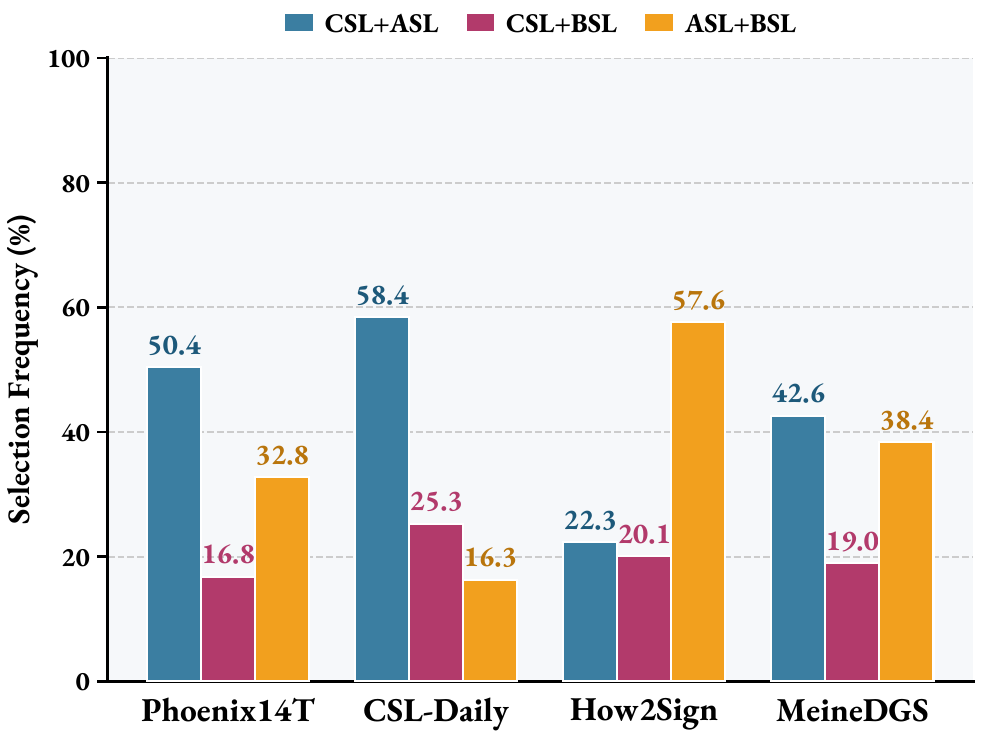}
    \caption{Per-dataset expert pair selection ($k{=}2$).}
    \label{fig:expert-selection}
  \end{minipage}
\end{figure}
\noindent\textbf{Expert Contribution Analysis.}
To understand how motion-level knowledge transfers across languages, we remove one expert at a time from the full three-backbone configuration and report the resulting BLEU-4 drop in Tab.~\ref{tab:leave-one-out}.
When the downstream dataset shares a language with a pretrained backbone, removing that backbone causes the largest performance drop. On CSL-Daily, removing the CSL expert leads to a substantial 6.56 BLEU-4 decrease, far exceeding the impact of removing ASL (1.48) or BSL (0.43). Similarly, on How2Sign, removing the ASL expert causes a 5.6-point drop, while the other removals have minimal effect (0.5 and 1.3). These results confirm that the framework effectively identifies and relies on the language-matched expert when one is available.
For the DGS datasets, where no matching backbone was seen during pretraining, the pattern is more nuanced. On Phoenix14T, the ASL expert is the most critical (4.74 drop), followed by BSL (3.76) and CSL (3.10), indicating that all three experts contribute meaningfully and that transfer to DGS draws on complementary cues rather than a single dominant source. On MeineDGS, a similar trend holds with ASL again being the most impactful (1.7 drop), while BSL and CSL contribute comparably (1.1 and 0.7). These drop patterns are consistent with the expert pair selection frequencies shown in Fig.~\ref{fig:expert-selection}. On CSL-Daily and How2Sign, the two pairs containing the language-matched backbone together account for 83.7\% (CSL+ASL and CSL+BSL) and 79.9\% (CSL+ASL and ASL+BSL) of selections, respectively, whereas on the DGS datasets the selection is more distributed, with no single pair exceeding 51\%.
The asymmetric drop patterns indicate language-specific rather than generic 
temporal representations, with expert importance shaped by linguistic 
relatedness and dataset properties. The language-matched backbone dominates 
when available, while for languages absent from Stage~I pretraining, the 
framework distributes reliance across experts, explaining the strong 
cross-lingual transfer in our main results. 

\begin{figure}[t]
\centering
\begin{minipage}[t]{0.48\linewidth}
    \null
    \centering
    \includegraphics[width=\linewidth]{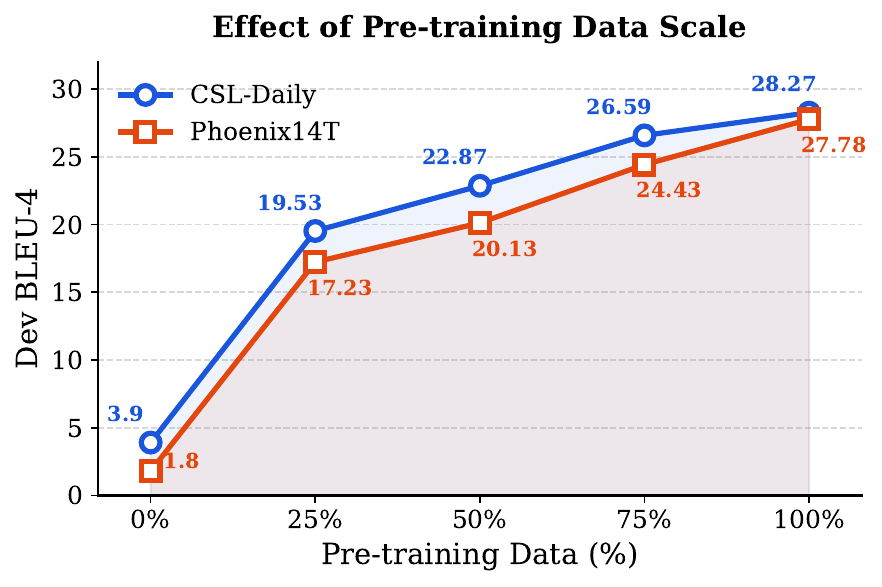}
    \caption{Pretraining Data Scale Impact.}
    \label{fig:scale}
\end{minipage}%
\hfill
\begin{minipage}[t]{0.50\linewidth}
    \null
    \centering
    \footnotesize
    \renewcommand{\arraystretch}{1.1}

    \captionof{table}{Fusion method comparison. BLEU-4 on Phoenix14T and CSL-Daily for combining the three pretrained experts via uniform averaging, feature concatenation, or our input-dependent adaptive gating. 
    % The best score on each dataset is highlighted in \textbf{bold}.
    }
    \label{tab:fusion}

    \resizebox{\linewidth}{!}{%
    \begin{tabular}{l|c|c}
    \toprule
    & \textbf{Phoenix14T} & \textbf{CSL-Daily} \\
    \cmidrule(lr){2-3}
    \textbf{Fusion Method} & B-4\textsuperscript{$\uparrow$} & B-4\textsuperscript{$\uparrow$} \\
    \midrule
    Average all experts ($\frac{1}{N}$) & 24.12 & 25.08 \\
    Concatenate experts & 24.30 & 25.75 \\
    \midrule
    \rowcolor[HTML]{e3f2fd}
    \textbf{SIGNET(adaptive gating)} & \textbf{27.78} & \textbf{28.27} \\
    \bottomrule
    \end{tabular}%
    }
\end{minipage}%
\end{figure}

\noindent\textbf{Impact of Pretraining Data Scale.} To examine how pretraining data volume affects downstream translation, we randomly subsample varying proportions (25\%, 50\%, 75\%, 100\%) from each of the three pretraining datasets (CSL, ASL, and BSL) for Stage~I. As shown in Fig.~\ref{fig:scale}, translation performance improves consistently as the pretraining data increases, and notably the trend does not plateau at full scale, indicating that the learned representations are not yet saturated and that the bottleneck lies in pretraining coverage rather than in the adaptation modules. This confirms that \textbf{SIGNET} benefits from larger-scale pretraining and suggests further gains as additional sign language data become available.

\noindent\textbf{SIGNET vs. Naive Combination.} As shown in Tab.~\ref{tab:fusion}, simply combining multiple pretrained backbones via concatenation or averaging assumes uniform relevance and increases overfitting risk. Averaging treats every expert as equally informative regardless of the input, diluting the language-matched backbone, while concatenation inflates the input dimensionality and provides more opportunities to overfit the limited downstream data. Both underperform. Averaging yields 24.12 and 25.08 BLEU-4 on Phoenix14T and CSL-Daily, respectively, while concatenation achieves 24.30 and 25.75. In contrast, our adaptive gating conditions the fusion on each input, sparsely activating only the most relevant experts, which preserves expert specialisation while limiting parameter growth. This substantially outperforms both static baselines (+3.5 BLEU-4 on average), showing that learned sparse routing effectively balances sparsity and diversity.

\section{Conclusion}
We presented \textbf{SIGNET}, a framework for motion-level 
knowledge transfer across sign languages that treats independently pretrained 
backbones as frozen visual experts with distinct motion-level specialisations.
Through hand-prior-driven routing and lightweight adaptation, \textbf{SIGNET} 
integrates multiple experts using only 1.5M trainable video encoder parameters 
while matching methods requiring orders of magnitude more compute. Our 
analysis reveals asymmetric expert reliance patterns consistent with 
language-specific rather than generic representations, with the framework 
automatically discovering complementary expertise for languages absent
from pretraining. Comprehensive experiments demonstrate state-of-the-art translation performance among gloss-free methods across multiple datasets, while also surpassing prior methods on sign language recognition.

\section*{Acknowledgements}
{\sloppy
This work was supported by EPSRC grant APP24554 (SignGPT-EP/Z535370/1), EPSRC grant APP78083 (UMCS UKRI3927) and through funding from Google.org via the AI for Global Goals scheme.
The authors acknowledge the use of Isambard-AI National AI Research Resource (AIRR) funded by UK DSIT via UKRI and STFC [ST/AIRR/I-A-I/1023].
This work reflects only the authors' views and the funders are not responsible for any use that may be made of the information it contains.\par}
\bibliographystyle{splncs04}
\bibliography{main}

\newpage
% =====================================================================
% Appendix / supplementary material body.
% Shared by main.tex (arXiv: appended inline) and supplementary.tex
% (camera-ready: compiled as its own document).
% =====================================================================
{
% \appendix switches section numbering to A, B, C, ... It also lets cleveref render
% \cref{...} of appendix sections as "Appendix A" (not "Sec. A"), and works with
% hyperref to give the appendix unique link anchors (no clashes with the main paper).
\appendix
% Force cleveref to name section-/subsection-counter labels as "Appendix" deterministically.
% (Relying on cleveref's automatic \appendix detection is version-dependent and renders
% "Sec. A" on some TeXLive versions, e.g. Overleaf.)
\crefalias{section}{appendix}
\crefalias{subsection}{subappendix}
% llncs places each paper at "chapter" level, so its \sections land at cleveref's
% "subappendix" depth. Name both levels "Appendix" so \cref renders "Appendix A"
% consistently across TeXLive versions (Overleaf included).
\crefname{appendix}{Appendix}{Appendices}\Crefname{appendix}{Appendix}{Appendices}
\crefname{subappendix}{Appendix}{Appendices}\Crefname{subappendix}{Appendix}{Appendices}
% Force UNIQUE hyperref anchors for appendix sections, independent of how a given
% hyperref version handles \appendix. Without this, appendix sections can reuse the
% main paper's anchors (section.1.1, ...) so \cref links jump to the wrong place
% (observed on Overleaf). These anchors are used for both the link target and the
% destination, so they stay consistent.
\renewcommand{\theHsection}{appendix.\arabic{section}}
\renewcommand{\theHsubsection}{appendix.\arabic{section}.\arabic{subsection}}
\setcounter{tocdepth}{2}

\begingroup
\centering
{\Large\bfseries SIGNET: Motion-Level Knowledge Transfer for\\
Cross-Language Sign Language Translation\par}
\vspace{10pt}
{\large\bfseries Appendix\par}
\endgroup

\vspace{1em}

\noindent This document provides additional technical details that complement the main paper, including details of the proposed sign language backbone (\cref{sec:backbone}), further implementation details (\cref{main:experiments}), qualitative results (\cref{main:results}), and additional experiments and ablations (\cref{main:ablation}).

\vspace{0.5em}

\etocsettocstyle{\section*{\contentsname}}{}
% Anchor the local TOC above section level so the appendix \sections (A, B, ...)
% are listed. Without this, etoc anchors at section level and only shows subsections.
\etocsetlocaltop{chapter}
% Black TOC entries only (default hyperref linkcolor is red); scoped so other
% appendix links keep their normal colour.
{\hypersetup{linkcolor=black}\localtableofcontents}
\newpage

% \setcounter{page}{1}

% % Manual title block - no \maketitle to avoid blank page and title in TOC
% \begingroup
% \centering
% {\large\bfseries Appendix\\[6pt]
% \textbf{SIGNET}: Motion-Level Knowledge Transfer for\\
% Cross-Language Sign Language Translation\par
% }

% % \textbf{SIGNET}: Motion-Level Knowledge Transfer for Cross-Language Sign Language Translation
% \endgroup
% \thispagestyle{empty}  % <-- suppress header on first supp page

% \noindent This document provides additional technical details that complement the main paper, including details of the proposed sign language backbone (\cref{sec:backbone}), further implementation details (\cref{main:experiments}), qualitative results (\cref{main:results}), and additional experiments and ablations (\cref{main:ablation}).

% % \makeatletter
% % \@starttoc{toc}
% % \makeatother
% \etoclocaltableofcontents
% \newpage

\section{Backbone Details}\label{sec:supp:backbone}
We provide comprehensive details on our sign language backbone architecture described in \cref{sec:backbone} of the main paper. This section first explains our pose extraction pipeline and regional partitioning strategy (Sec.~\ref{supp:pose_extraction}), followed by detailed architectural specifications of the ST-GCN (Sec.~\ref{supp:stgcn_details}).
\begin{figure}
    \centering
    \setlength{\belowcaptionskip}{-25pt}
    \includegraphics[width=1\linewidth]{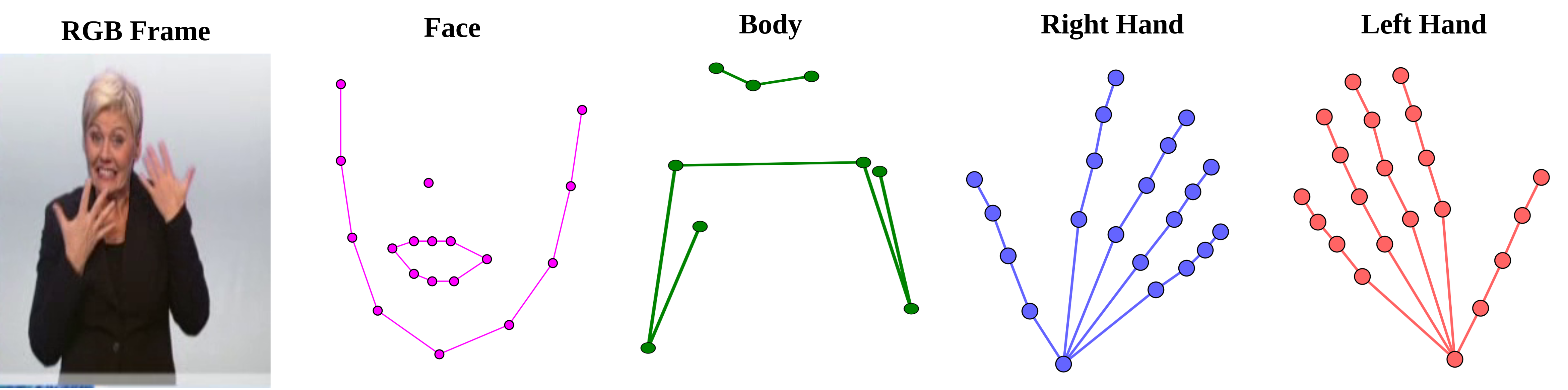}
    \caption{Visualisation of 2D keypoints extracted.}
    \label{fig:placeholder}
\end{figure}
\subsection{Pose Extraction and Region Partitioning}
\label{supp:pose_extraction}
\noindent\textbf{Whole-Body Keypoint Extraction.}
We utilise RTMPose-x~\cite{jiang2023rtmpose} from the MMPose framework to extract comprehensive whole-body pose information from sign language videos. RTMPose-x provides 133 keypoints covering the entire body, including fine-grained hand and facial landmarks essential for sign language understanding.\\

\noindent\textbf{Anatomical Region Partitioning.}
As described in Sec.~\ref{sec:preprocess} of the main paper, we partition the extracted keypoints into four anatomical regions. The specific keypoint indices for each region are:
\begin{itemize}
    \item \textbf{Body}: Indices $\{1, 4-11\}$ ($J_b = 9$ keypoints)
    \item \textbf{Left Hand}: Indices $\{92-112\}$ ($J_{lh} = 21$ keypoints)
    \item \textbf{Right Hand}: Indices $\{113-133\}$ ($J_{rh} = 21$ keypoints)
    \item \textbf{Face}: Indices $\{24, 26, 28, 30, 32, 34, 36, 38, 40, 54, 84-91\}$ ($J_f = 18$ keypoints)
\end{itemize}
\noindent\textbf{Root-Relative Normalisation.}
To ensure translation invariance and focus on relative motion patterns, we apply root-relative normalisation for the hands and face regions. Specifically, we designate keypoint indices 92, 113, and 54 as root nodes for the left hand, right hand, and face, respectively. Each keypoint in these regions is normalised relative to its corresponding root:
\begin{equation}
\mathcal{P}_{r}^{\text{norm}}(j) = \mathcal{P}_{r}(j) - \mathcal{P}_{r}(\text{root}), \quad j \in \mathcal{J}_r
\end{equation}
where $\mathcal{J}_r$ denotes the set of keypoint indices for region $r$. The body region is not subjected to root-relative normalisation, preserving absolute spatial positioning information that captures global body orientation and location.\\
\vspace{-20pt}
\subsection{Input Representation to ST-GCN Modules}
The 2D keypoint sequences $\mathcal{P}_{r} \in \mathbb{R}^{T \times J_{r} \times 2}$ described in the main paper are extended with confidence scores from RTMPose before being fed into the ST-GCN modules. Specifically, each keypoint $(x, y)$ is extended to $(x, y, c)$, where $c \in [0, 1]$ represents the detection confidence. The extended input therefore has shape $\mathbb{R}^{T \times J_{r} \times 3}$.
Each ST-GCN module processes this 3-channel representation through a projection layer that maps $\mathbb{R}^{3} \rightarrow \mathbb{R}^{64}$ before graph convolution. The confidence scores provide the network with information about keypoint detection reliability, which is particularly valuable for handling occlusions and detection uncertainty in sign language videos.\\

\begin{table*}[t]
\setlength{\belowcaptionskip}{-15pt}
\centering
\scriptsize
\begin{minipage}[b]{0.48\linewidth}
\centering
\renewcommand{\arraystretch}{1.3}  % <-- add here
\resizebox{\linewidth}{!}{
\begin{tabular}{l|l|c|c}
\hline
\textbf{Stage} & \textbf{Layer Type} & \textbf{Channels} & \textbf{Temporal Kernel} \\
\hline
\rowcolor{gray!30}
\multicolumn{4}{c}{\textbf{Spatial GCN}} \\
\hline
Stage 1 & GCN + TCN & $64 \rightarrow 64$ & 1 \\
Stage 2 & GCN + TCN & $64 \rightarrow 128$ & 1 \\
Stage 3 & GCN + TCN & $128 \rightarrow 256$ & 1 \\
\hline
\rowcolor{gray!30}
\multicolumn{4}{c}{\textbf{Temporal GCN}} \\
\hline
Stage 1 & GCN + TCN & $256 \rightarrow 256$ & 5 \\
Stage 2 & GCN + TCN & $256 \rightarrow 256$ & 5 \\
Stage 3 & GCN + TCN & $256 \rightarrow 256$ & 5 \\
\hline
\end{tabular}}
\caption{ST-GCN Detailed Architecture. GCN: Graph Convolutional Network. TCN: Temporal Convolutional Network. All stages include residual connections, BN, ReLU, Dropout. Spatial kernel size: 2 partitions (distance-based graph).}
\label{tab:stgcn_detailed}
\end{minipage}
\hfill
\begin{minipage}[b]{0.51\linewidth}
\centering
\renewcommand{\arraystretch}{1.2}  % <-- add here
\scriptsize
\resizebox{\linewidth}{!}{
\begin{tabular}{c|c|c|c|c|c}
\hline
\textbf{Stream} & \textbf{Component} & \textbf{Input} & \textbf{Output} & \textbf{Kernel} & \textbf{Graph} \\
\hline
\multirow{3}{*}{\makecell{\textbf{Body} \\[1ex] \scalebox{1.2}{$\displaystyle (\mathcal{G}_{b})$}}} & \textbf{Projection} & $ T \times J_b \times 3$ & $T \times J_b \times 64$ & - & - \\
& \textbf{Spatial GCN} & $T \times J_b \times 64$ & $T \times J_b \times 256$ & $(1, 2)$ & Distance \\
& \textbf{Temporal GCN} & $T \times J_b \times 256$ & $T \times J_b \times 256$ & $(5, 2)$ & Distance \\
& \textbf{Mean Pool} & $T \times J_b \times 256$ & $T \times 256$ & - & - \\
\hline
\multirow{3}{*}{\makecell{\textbf{Right Hand} \\[1ex] \scalebox{1.2}{$\displaystyle (\mathcal{G}_{rh})$}}} & \textbf{Projection} & $ T \times J_{rh} \times 3$ & $T \times J_{rh} \times 64$ & - & - \\
& \textbf{Spatial GCN} & $T \times J_{rh} \times 64$ & $T \times J_{rh} \times 256$ & $(1, 2)$ & Distance \\
& \textbf{Temporal GCN} & $T \times J_{rh} \times 256$ & $T \times J_{rh} \times 256$ & $(5, 2)$ & Distance \\
& \textbf{Mean Pool} & $T \times J_{rh} \times 256$ & $T \times 256$ & - & - \\
\hline
\multirow{3}{*}{\makecell{\textbf{Left Hand} \\[1ex] \scalebox{1.2}{$(\mathcal{G}_{lh})$}}} & \textbf{Projection} & $ T \times J_{lh} \times 3$ & $T \times J_{lh} \times 64$ & - & - \\
& \textbf{Spatial GCN} & $T \times J_{lh} \times 64$ & $T \times J_{lh} \times 256$ & $(1, 2)$ & Distance \\
& \textbf{Temporal GCN} & $T \times J_{lh} \times 256$ & $T \times J_{lh} \times 256$ & $(5, 2)$ & Distance \\
& \textbf{Mean Pool} & $T \times J_{lh} \times 256$ & $T \times 256$ & - & - \\
\hline
\multirow{3}{*}{\makecell{\textbf{Face} \\[1ex] \scalebox{1.2}{$(\mathcal{G}_{f})$}}} & \textbf{Projection} & $ T \times J_{f} \times 3$ & $T \times J_{f} \times 64$ & - & - \\
& \textbf{Spatial GCN} & $T \times J_{f} \times 64$ & $T \times J_{f} \times 256$ & $(1, 2)$ & Distance \\
& \textbf{Temporal GCN} & $T \times J_{f} \times 256$ & $T \times J_{f} \times 256$ & $(5, 2)$ & Distance \\
& \textbf{Mean Pool} & $T \times J_{f} \times 256$ & $T \times 256$ & - & - \\
\hline
\multicolumn{2}{l|}{\textbf{Fusion (Concat)}} & \multicolumn{4}{c}{$T \times 1024$} \\ \hline
\multicolumn{2}{l|}{\textbf{Pose Projection}} & $T \times 1024$ & $T \times 768$ & - & - \\
\hline
\multicolumn{2}{l|}{\textbf{Number of Parameters}} & \multicolumn{4}{c}{5.3 M} \\ \hline
\end{tabular}}
\caption{Sign Language Backbone Architecture. $T$: Temporal length, $J_b$: Body keypoints, $J_{rh}$: Right Hand keypoints, $J_{lh}$: Left Hand keypoints, $J_f$: Face keypoints, Kernel format: (temporal size, spatial partitions).}
\label{tab:stgcn_architecture}
\end{minipage}
\end{table*}

\noindent\textbf{Robustness to Keypoint Detection Noise.}
Our pipeline leverages RTMPose-x~\cite{jiang2023rtmpose}, a state-of-the-art whole-body pose estimator that has demonstrated strong generalisation across diverse domains and recording conditions. RTMPose itself provides a robust foundation for keypoint extraction; however, no pose estimator is immune to detection errors under challenging scenarios such as heavy occlusions, extreme motion blur, or unusual camera viewpoints. To further mitigate these residual detection failures, we apply a confidence-based filtering step prior to feeding keypoints into the ST-GCN modules: any keypoint with a detection confidence below $\tau = 0.3$ is zeroed out, i.e., $(x, y, c) \rightarrow (0, 0, 0)$. This hard thresholding removes spurious or poorly localised detections that survive the pose estimator.

This filtering acts as implicit data augmentation, forcing the model to learn from partially incomplete skeletons, while the continuous confidence values allow the network to soft-weight keypoint reliability during graph reasoning. Together with the inherent robustness of RTMPose as the upstream detector, this creates a three-level robustness mechanism that enables \textbf{SIGNET} to generalise across heterogeneous recording environments.

\subsection{Architecture Details}
\label{supp:stgcn_details}

We provide detailed architectural specifications for our ST-GCN modules. Tab.~\ref{tab:stgcn_detailed} presents the internal structure of the two processing stages, spatial and temporal, employed within each regional ST-GCN stream. Tab.~\ref{tab:stgcn_architecture} provides a comprehensive view of the complete multi-stream backbone, showing how the four regional streams are processed in parallel and subsequently fused.\\

\noindent\textbf{Spatial and Temporal GCN.}
Each regional ST-GCN stream ($\mathcal{G}_r$) consists of two sequential processing stages. The Spatial GCN (Tab.~\ref{tab:stgcn_detailed}, top) progressively expands the feature dimensionality from 64 to 256 channels through three stages with temporal kernel size 1, focusing on modelling spatial relationships between joints within each frame. The Temporal GCN (Tab.~\ref{tab:stgcn_detailed}, bottom) maintains 256 channels across three stages with temporal kernel size 5, capturing motion dynamics and temporal dependencies across frames. All stages employ adaptive graph learning with 2 spatial partitions based on distance-based connectivity, along with residual connections, batch normalisation, ReLU activation, and dropout regularisation.\\

\noindent\textbf{Multi-Stream Architecture.}
Tab.~\ref{tab:stgcn_architecture} illustrates how each of the four anatomical regions is processed through identical ST-GCN structures. After independent processing, each stream produces a $T \times 256$ dimensional feature representation via mean pooling over the spatial (joint) dimension. These four regional features are concatenated to form a unified $T \times 1024$ representation, which is then projected to $T \times 768$ dimensions for alignment with text encoders. Note that the left and right hand streams share identical network parameters to reduce model complexity while maintaining representational capacity.

\section{Implementation Details}\label{sec:supp:impl}
\noindent\textbf{Datasets.}
Tab.~\ref{tab:datasets} lists all datasets employed in this study, reporting their source, duration, and research community popularity.\\
\begin{table*}[t]
\setlength{\belowcaptionskip}{-15pt}
\centering
\begin{minipage}[b]{0.45\linewidth}
\centering
\scriptsize
\renewcommand{\arraystretch}{1.3}
\resizebox{\linewidth}{!}{
\begin{tabular}{l|c|c|c}
\hline
\textbf{Dataset} & \textbf{Hours} & \textbf{Text Vocab.} & \textbf{Glosses}\\
\hline
BSL Corpus~\cite{schembri2013building} & 125 & \textbf{--} & 1,800\\
Phoenix14T~\cite{nslt} & 11 & 3K & 1,066\\
MeineDGS~\cite{meinedgs_3} & 50 & 18K & 10,000 \\
CSL-Daily~\cite{csldaily} & 23 & 2K & 2,000\\
How2Sign~\cite{how2sign} & 79 & 16K & N/A\\
\makecell[l]{YouTube-ASL~\cite{youtube_asl}\\(YT-ASL)} & 984 & 60K & N/A \\
\makecell[l]{YouTube-SL-25~\cite{youtube_sl}\\(YT-SL25)} & 3,207 & \textbf{--} & N/A \\
CSL-News~\cite{li2025uni} & 1,985 & 5K & N/A \\
BOBSL~\cite{bobsl} & 1,467 & 77K & N/A \\
\hline
\end{tabular}}
\caption{Sign Language Translation Datasets Overview. N/A demonstrates that gloss annotations are not available for the dataset, \textbf{--} indicates that the value was not reported.}
\label{tab:datasets}
\end{minipage}
\hfill
\begin{minipage}[b]{0.54\linewidth}
\centering
\scriptsize
\renewcommand{\arraystretch}{1.3}
\resizebox{\linewidth}{!}{
\begin{tabular}{l|ccc}
\hline
\textbf{Parameter} & \textbf{Stage~I} & \textbf{Stage~II} & \textbf{Stage~III} \\
\hline
Label smoothing & 0.2 & \ding{55} & 0.2 \\
Number of beams & 4 & \ding{55} & 4 \\
Optimiser & \multicolumn{3}{c}{AdamW} \\
Optimiser Momentum ($\beta_1/\beta_2$) & \multicolumn{3}{c}{0.9/0.999} \\
Schedule & \multicolumn{3}{c}{Cosine} \\
Weight decay & \multicolumn{3}{c}{$1 \times 10^{-4}$} \\
Batch size & \multicolumn{3}{c}{8} \\
Gradient accumulation & \multicolumn{3}{c}{4} \\
Learning rate & \multicolumn{3}{c}{$3 \times 10^{-4}$} \\
Minimum learning rate & \multicolumn{3}{c}{$1 \times 10^{-7}$} \\
Epochs & 20 & 20 & 40 \\
Warmup epochs & 0 & 2 & 4 \\
Router Learning Rate & \ding{55} & $8 \times 10^{-4}$ & $8 \times 10^{-4}$ \\
$\tau_{\text{Gumbel-Softmax}}$ & \ding{55} & 1.0 & 1.0 \\
$\tau_{\text{Contrastive}}$ & \ding{55} & 0.07 & \ding{55} \\
\hline
\end{tabular}}
\caption{Training Hyperparameters.}
\label{tab:hyperparameters}
\end{minipage}
\end{table*}

\noindent\textbf{Data Augmentation.}
During training, we apply two augmentation strategies to improve generalisation. First, we add Gaussian noise to the joint coordinates with $\sigma = 0.01$, which encourages robustness to minor variations in pose estimation. Second, we randomly drop 15\% of input frames along the temporal axis, simulating natural variations in signing speed and forcing the model to handle incomplete sequences.\\

\noindent\textbf{LLM Encoder-Decoder.}
We use mT5-base~\cite{mt5}, a 580M-parameter multilingual encoder-decoder model, for text generation. The model consists of 12-layer encoder and decoder Transformers with hidden dimension 768, 12 attention heads, and feed-forward dimension 2048. mT5 uses a SentencePiece tokeniser with 250K vocabulary, pretrained on the mC4 corpus covering 101 languages. 
Visual features from the sign language backbone (Sec.~\ref{supp:stgcn_details}) are projected to match the 768-dimensional LLM hidden space and fed to the decoder via cross-attention. We initialise from pretrained mT5-base checkpoints and apply stage-specific training strategies.\\

\noindent\textbf{Stage I: Pretraining on Large Datasets.}
For each large-scale dataset, we train a sign language backbone paired with its corresponding LLM using the configuration specified in Tab.~\ref{tab:hyperparameters}. We save the pretraining weights corresponding to the highest BLEU-4 score achieved on the validation set of each dataset.\\

\noindent\textbf{Lightweight Expert Adapters.}
We employ lightweight LoRA-style adapters~\cite{hu2022lora} to efficiently adapt frozen expert outputs. Each adapter implements a low-rank residual transformation with hidden dimension $H=768$, rank $R=16$, and dropout rate $0.1$:
\begin{equation}
\tilde{\mathcal{Z}}^{i} = \mathcal{Z}^{i} + 
\operatorname{Dropout}(\mathbf{B}_{i}\big(\mathbf{A}_{i}(\operatorname{LayerNorm}(\mathcal{Z}^{i}))\big)),
\end{equation}
where $\mathbf{A}_{i} \in \mathbb{R}^{H \times R}$ and $\mathbf{B}_{i} \in \mathbb{R}^{R \times H}$. We initialise $\mathbf{A} \sim \mathcal{N}(0, 10^{-4})$ and $\mathbf{B} = \mathbf{0}$ to preserve pretrained features initially.\\

\noindent\textbf{Feature Aggregation.}
Following Tab.~\ref{tab:stgcn_architecture}, we extract hand-specific features from each sign-specialised expert by mean-pooling the temporal dimension and concatenating left and right hand representations:
\begin{equation}
\mathbf{H}_{i} = [\text{Mean}_{T}(\mathcal{F}_{\text{rh}}^{i}); \text{Mean}_{T}(\mathcal{F}_{\text{lh}}^{i})] \in \mathbb{R}^{B \times D},
\end{equation}
where $B$ is the batch size, $D=512$, and $\mathcal{F}_{\text{rh}}^{i}, \mathcal{F}_{\text{lh}}^{i}$ denote the right and left hand features from expert $i$, respectively. As shown in Alg.~\ref{alg:feature_aggregation} (line 3), we generate a data-dependent query representation using a two-layer MLP with hidden dimension 512:
\begin{equation}
\mathbf{q} = \text{MLP}(\bar{\mathbf{H}}) = \mathbf{W}_2 \cdot \text{ReLU}(\mathbf{W}_1 \cdot \bar{\mathbf{H}}),
\end{equation}
where $\mathbf{W}_1, \mathbf{W}_2 \in \mathbb{R}^{512 \times 512}$.\\

\noindent\textbf{Frozen Text Encoder.} We employ a separate frozen mT5-base encoder to extract reference text embeddings for contrastive learning in Stage II. This encoder shares the same architecture (12 layers, hidden dimension 768, 12 attention heads) and is initialised from pretrained mT5-base checkpoints.\\

\noindent\textbf{Stage II: Downstream Pretraining.}
In Stage II, we perform contrastive pretraining on each downstream dataset using InfoNCE loss. We train the model according to the hyperparameters listed in Tab.~\ref{tab:hyperparameters}. For each dataset, we save the model checkpoint that achieves the lowest validation InfoNCE loss.\\

\noindent\textbf{Stage III: Downstream Fine-tuning.}
For each downstream dataset, we fine-tune the Stage II checkpoint using the configuration specified in Tab.~\ref{tab:hyperparameters}. We save the checkpoint corresponding to the highest BLEU-4 score achieved on the validation set of each dataset.

\section{Additional Ablations}\label{sec:sup:ablation}

\subsection{Sensitivity to LLM Initialisation}
The source-LLM choice affects only datasets absent from Stage~I pretraining; languages seen during pretraining use their matching encoder--decoder. To verify the impact is negligible, we evaluate the two unseen-language datasets (Phoenix14T and MeineDGS), initialising the encoder--decoder from each pretrained source with three seeds each (\cref{tab:llm_sensitivity}). The cross-source spread is $\le$0.07 B-4 on Phoenix14T and $\le$0.06 on MeineDGS, with within-source standard deviation $\le$0.13 --- all within seed noise. We therefore initialise from the CSL model in the main paper, making the reported numbers conservative.

\begin{table}[h]
\centering
\footnotesize
\renewcommand{\arraystretch}{1.1}
% Measure the table into a box so the caption (placed below) matches its width.
\sbox0{%
\begin{tabular}{lcc}
\toprule
\textbf{Source LLM} & \textbf{Phoenix14T} (Test B-4) & \textbf{MeineDGS} (Test B-4) \\
\midrule
Init.\ from CSL & 27.82 $\pm$ 0.10 & 2.75 $\pm$ 0.07 \\
Init.\ from ASL & 27.89 $\pm$ 0.12 & 2.79 $\pm$ 0.07 \\
Init.\ from BSL & 27.85 $\pm$ 0.13 & 2.81 $\pm$ 0.10 \\
\bottomrule
\end{tabular}}%
\begin{minipage}{\wd0}
\centering
\usebox0
\caption{\textbf{Sensitivity to the source LLM} used to initialise the encoder--decoder for languages unseen in Stage~I (mean\,$\pm$\,std over three seeds, Test split). Differences are within seed noise; the main paper reports the CSL-initialised model.}
\label{tab:llm_sensitivity}
\end{minipage}
\end{table}
\newpage

\section{Qualitative Results}\label{sec:supp:qual}
\subsection{CSL-Daily}
Fig.~\ref{qual:csl} shows representative translation examples comparing \textbf{SIGNET} with two previous state-of-the-art methods. Our approach consistently generates more accurate and complete translations, better preserving spatial relationships and semantic content. For this comparison, we use examples from the supplementary material of Geo-Sign~\cite{fish2025geo} and re-implement Uni-Sign~\cite{li2025uni}, performing inference using their available weights on Hugging Face.
\begin{figure}[H]
    \centering
    \includegraphics[width=1\linewidth]{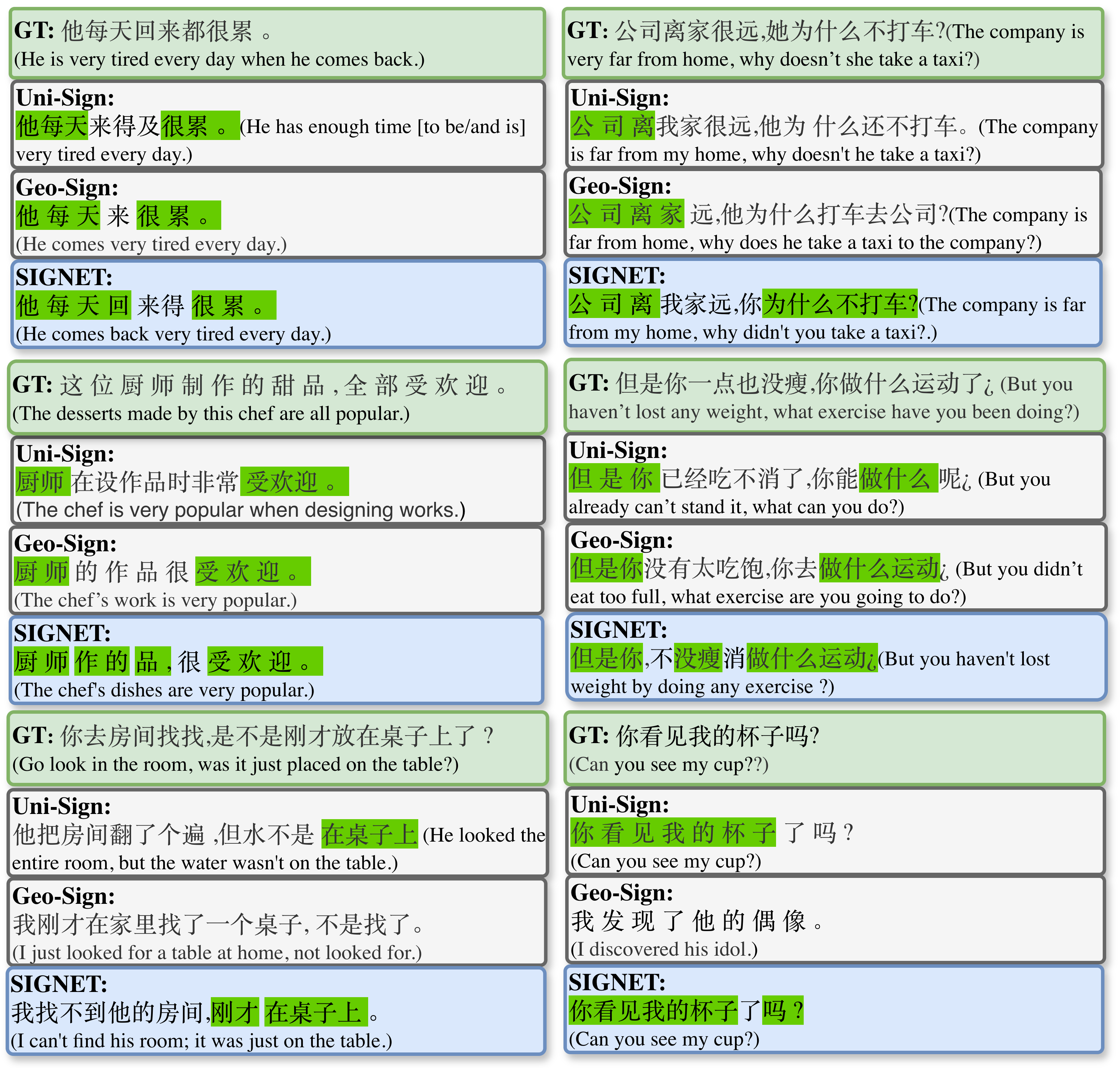}
    \caption{Qualitative results on CSL-Daily test set. {\setlength{\fboxsep}{2pt}\colorbox[HTML]{66cc00}{Green}} text indicates correct predictions matching the Ground Truth (\textbf{GT}).}
    \label{qual:csl}
\end{figure}
\subsection{How2Sign}
Fig.~\ref{qual:h2s} presents qualitative comparisons with SSVP-SLT-LSP~\cite{rust2024towards} and PGG-SLT~\cite{guo2025bridging}. Our results demonstrate translation quality comparable to SSVP-SLT-LSP while outperforming PGG-SLT in semantic accuracy and completeness. These qualitative observations align with our quantitative evaluation on the How2Sign benchmark. For this comparison, we use examples from the supplementary materials of SSVP-SLT-LSP and PGG-SLT.

\begin{figure}
    \centering
    \includegraphics[width=1\linewidth]{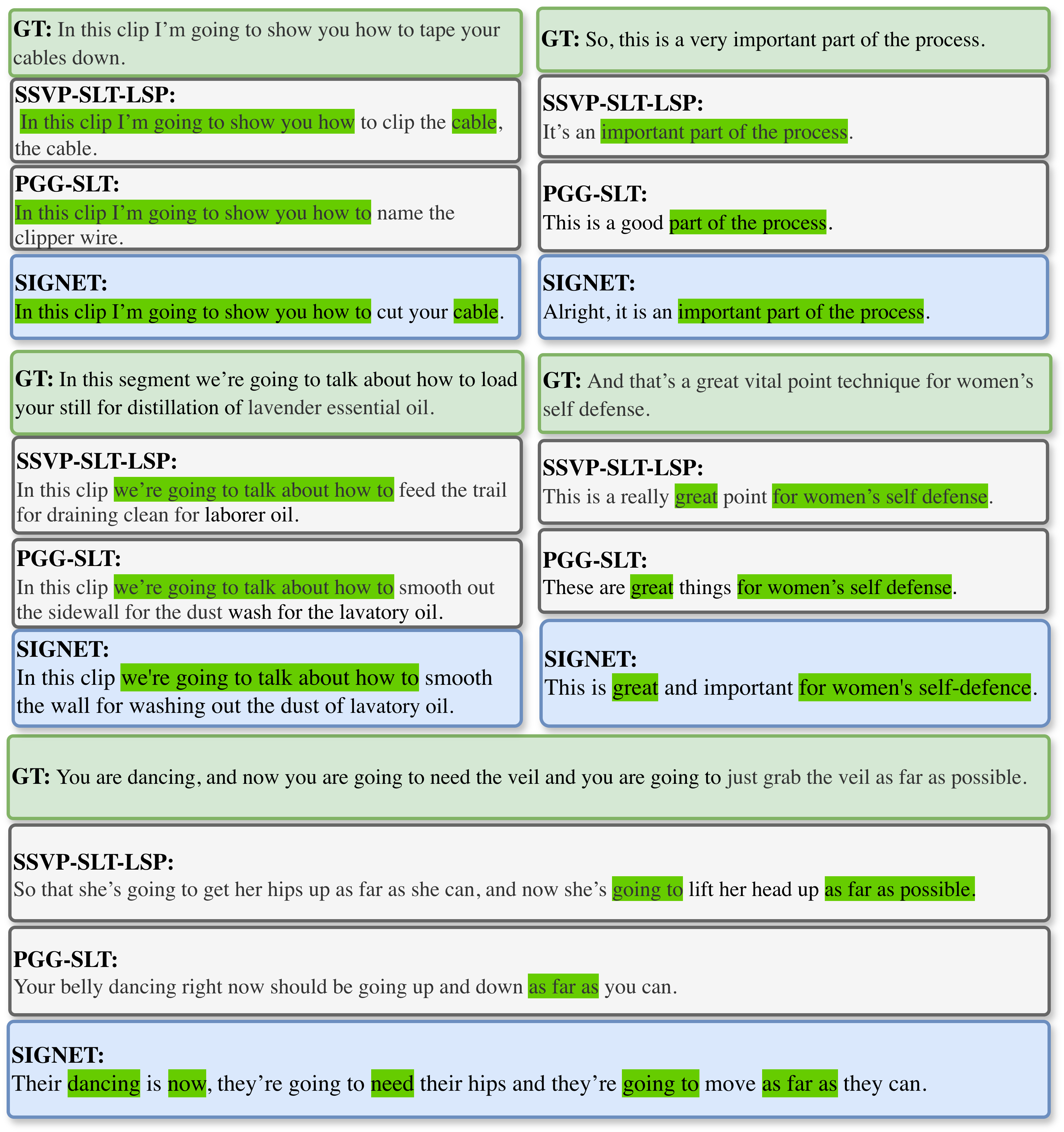}
    \caption{Qualitative results on How2Sign test set. {\setlength{\fboxsep}{2pt}\colorbox[HTML]{66cc00}{Green}} text indicates correct predictions matching the Ground Truth (\textbf{GT}).}
    \label{qual:h2s}
\end{figure}

\subsection{Phoenix14T}
Fig.~\ref{qual:phnx} shows qualitative comparisons with PGG-SLT~\cite{guo2025bridging} on the Phoenix14T dataset. PGG-SLT employs LLM-based instruction tuning with few-shot examples to reorder glosses before translation. 
\textbf{SIGNET} demonstrates superior translation quality, especially for longer sequences.
We use examples from the PGG-SLT supplementary material.
\begin{figure}[H]
    \centering
    \includegraphics[width=1\linewidth]{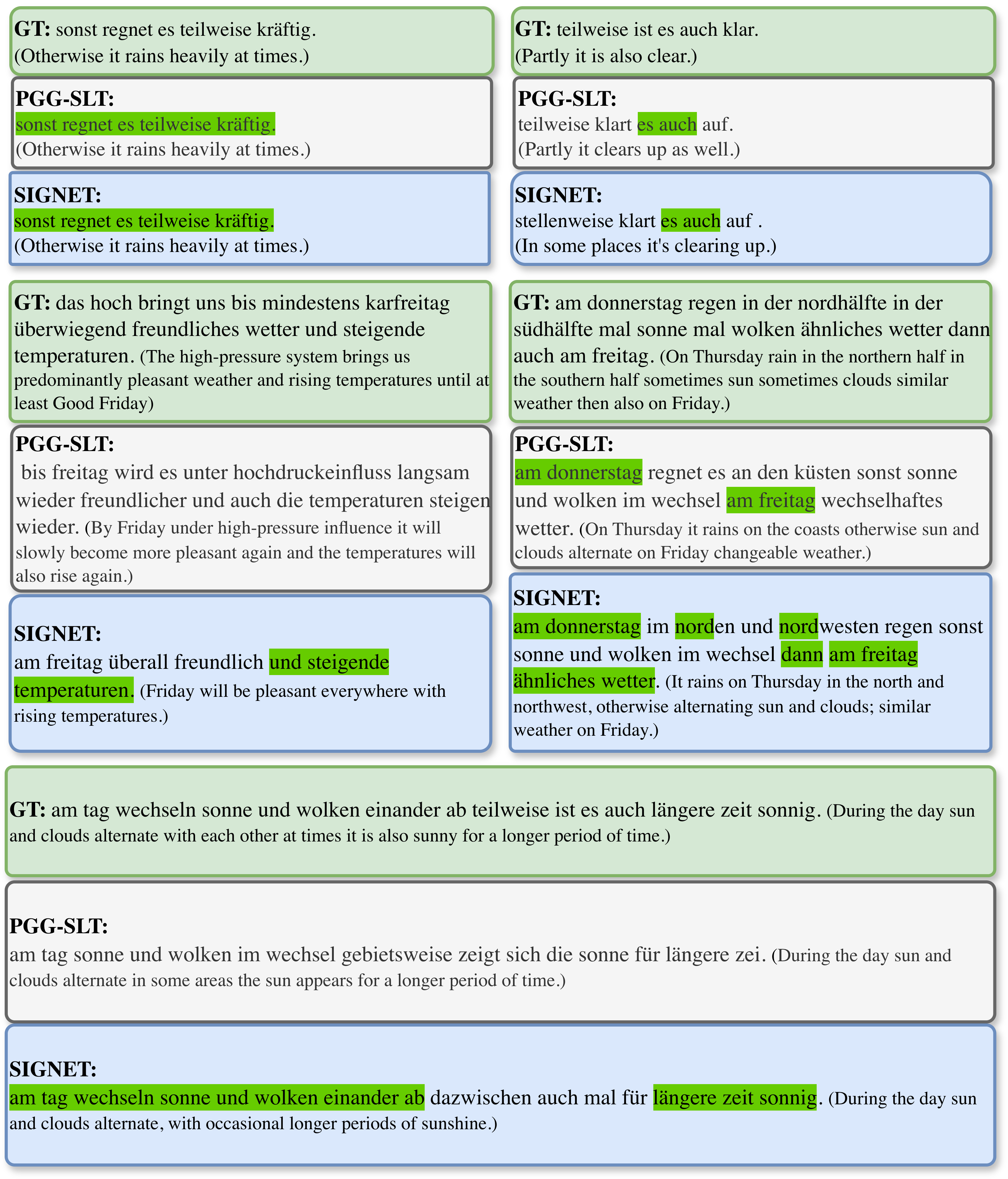}
    \caption{Qualitative results on Phoenix14T. {\setlength{\fboxsep}{2pt}\colorbox[HTML]{66cc00}{Green}} text indicates correct predictions matching the Ground Truth (\textbf{GT}).}
    \label{qual:phnx}
\end{figure}

\subsection{MeineDGS}
MeineDGS provides diverse data formats ranging from story-retelling to free-flowing conversation and is less frequently studied in prior work. As shown in Tab.~\ref{tab:datasets}, the dataset contains approximately 10{,}000 unique glosses within only 50 hours of video, resulting in a considerably higher gloss-to-duration ratio than other benchmarks. This lexical diversity makes MeineDGS particularly challenging, which explains the lower results compared to other datasets. \textbf{SIGNET} outperforms the previous best method, Sincan et al.~\cite{sincan2025spotter+}, which relies on a two-stage pipeline combining a sign spotter with GPT for translation. 
Fig.~\ref{qual:mdgs} presents qualitative comparisons with Sincan et al.~\cite{sincan2025spotter+}.
For this comparison, we use the three available qualitative examples from Sincan et al.~\cite{sincan2025spotter+} on the same MeineDGS split used in our experiments.
\begin{figure}[H]
    \centering
    \includegraphics[width=1\linewidth]{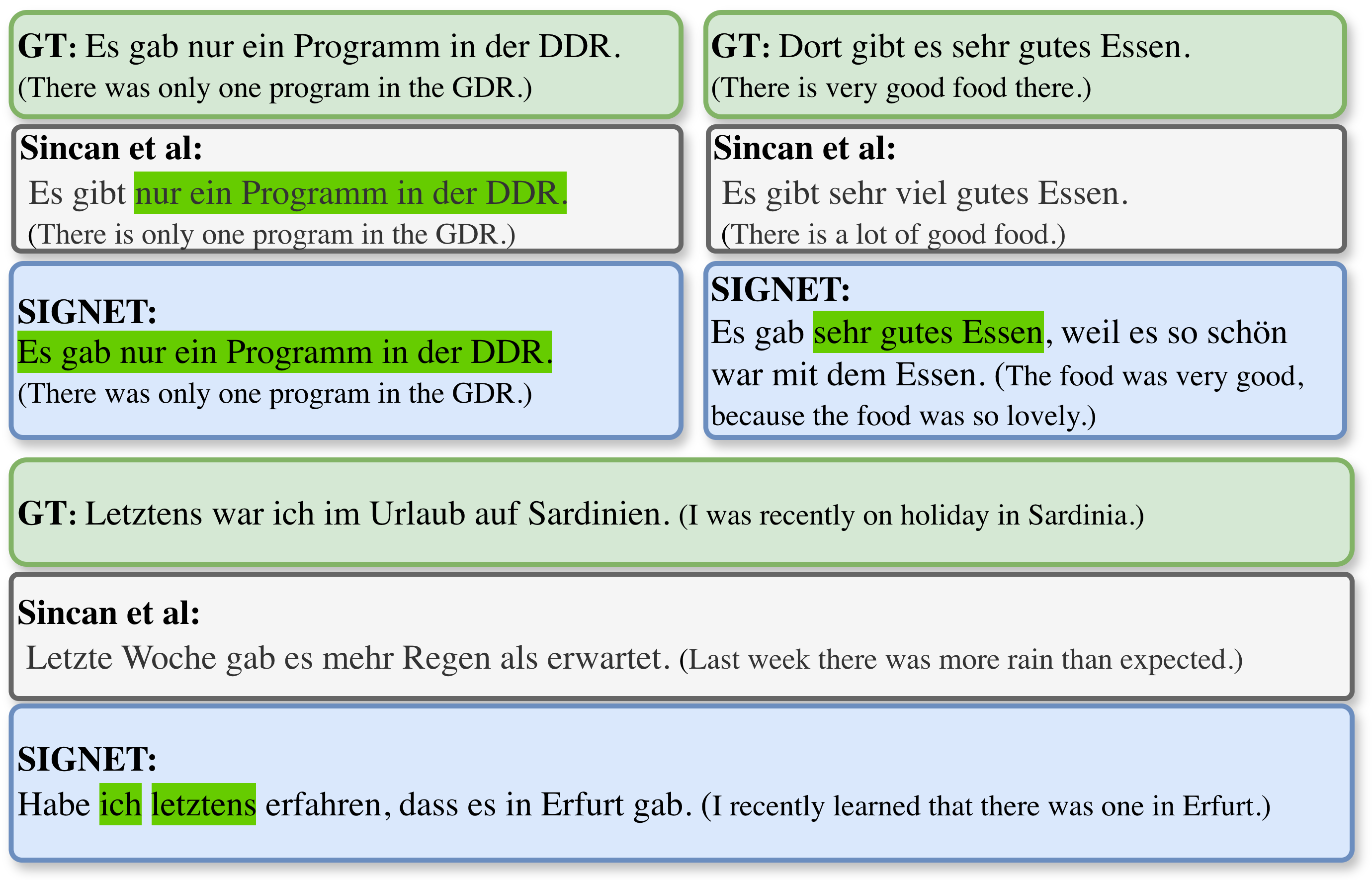}
    \caption{Qualitative results on MeineDGS test set. {\setlength{\fboxsep}{2pt}\colorbox[HTML]{66cc00}{Green}} text indicates correct predictions matching the Ground Truth (\textbf{GT}).}
    \label{qual:mdgs}
\end{figure}

}

\end{document}